\documentclass{article}
\usepackage[final]{corl_2022}
\usepackage{lipsum}
\usepackage{graphics,graphicx,caption,float,subcaption,booktabs,xcolor,multirow,array,color,ifthen,tabu,colortbl,dblfloatfix,url,xparse,mathtools,patchcmd,algorithm,algorithmic,amssymb,xspace,nicefrac,microtype,amsmath,amsthm,amsfonts,bm,ragged2e,tikz,stackengine,etoolbox,xpatch,enumerate,xstring,setspace,tabularx,makecell,changepage,cuted,titlesec,enumitem,wrapfig,tcolorbox, wrapfig, capt-of}
\usepackage{graphics,graphicx,float,booktabs,xcolor,multirow,array,color,ifthen,tabu,colortbl,dblfloatfix,url,xparse,mathtools,patchcmd,algorithm,algorithmic,amssymb,xspace,nicefrac,microtype,amsmath,amsthm,amsfonts,bm,ragged2e,tikz,stackengine,etoolbox,xpatch,enumerate,xstring,setspace,makecell,changepage,cuted,titlesec,enumitem,wrapfig,tcolorbox, wrapfig}
\usepackage{wrapfig}
\usepackage{capt-of}
\hypersetup{bookmarksopen,bookmarksnumbered,
pdfpagemode=UseOutlines,
colorlinks=true,
linkcolor=blue,
anchorcolor=blue,
citecolor=blue,
filecolor=blue,
menucolor=blue,
urlcolor=blue
}
\usepackage[capitalise,noabbrev,nameinlink]{cleveref}
\usepackage[english]{babel}

\title{VideoDex: Learning Dexterity from Internet Videos}
\author{Kenneth Shaw$^*$ $\qquad$ Shikhar Bahl$^*$ $\qquad$Deepak Pathak\\ \\Carnegie Mellon University}

\begin{document}
\newcommand{\our}{VideoDex\xspace}
\newcommand{\ours}{VideoDex\xspace}
\maketitle
\vspace{-0.20in}

\begin{abstract}
To build general robotic agents that can operate in many environments, it is often imperative for the robot to collect experience in the real world.  However, this is often not feasible due to safety, time, and hardware restrictions.  We thus propose leveraging the next best thing as real-world experience: internet videos of humans using their hands.  Visual priors, such as visual features, are often learned from videos, but we believe that more information from videos can be utilized as a stronger prior.  We build a learning algorithm, \ours, that leverages \textit{visual}, \textit{action}, and \textit{physical} priors from human video datasets to guide robot behavior.  These actions and physical priors in the neural network dictate the typical human behavior for a particular robot task.   We test our approach on a robot arm and dexterous hand-based system and show strong results on various manipulation tasks, outperforming various state-of-the-art methods. For videos and supplemental material visit our website at \url{https://video-dex.github.io}

\end{abstract}
\renewcommand{\thefootnote}{\fnsymbol{footnote}}
\footnotetext{$^*$Equal contribution, order decided by coin flip.}
\vspace{-0.12in}
\section{Introduction}
\label{sec:intro}
\vspace{-0.15in}

The long-standing dream of many roboticists is to see robots autonomously perform diverse tasks in diverse environments. 
To build a robot that can operate anywhere, many methods rely on successful robotic interaction data to train on.
However, deploying inexperienced, real-world robots to collect experience may require constant supervision which is infeasible. This poses a chicken-and-egg problem for robot learning because to collect experience safely, the robot already needs to be experienced. How do we get around this deadlock?

\begin{figure}[t!]
 \centering
 \includegraphics[width=\linewidth]{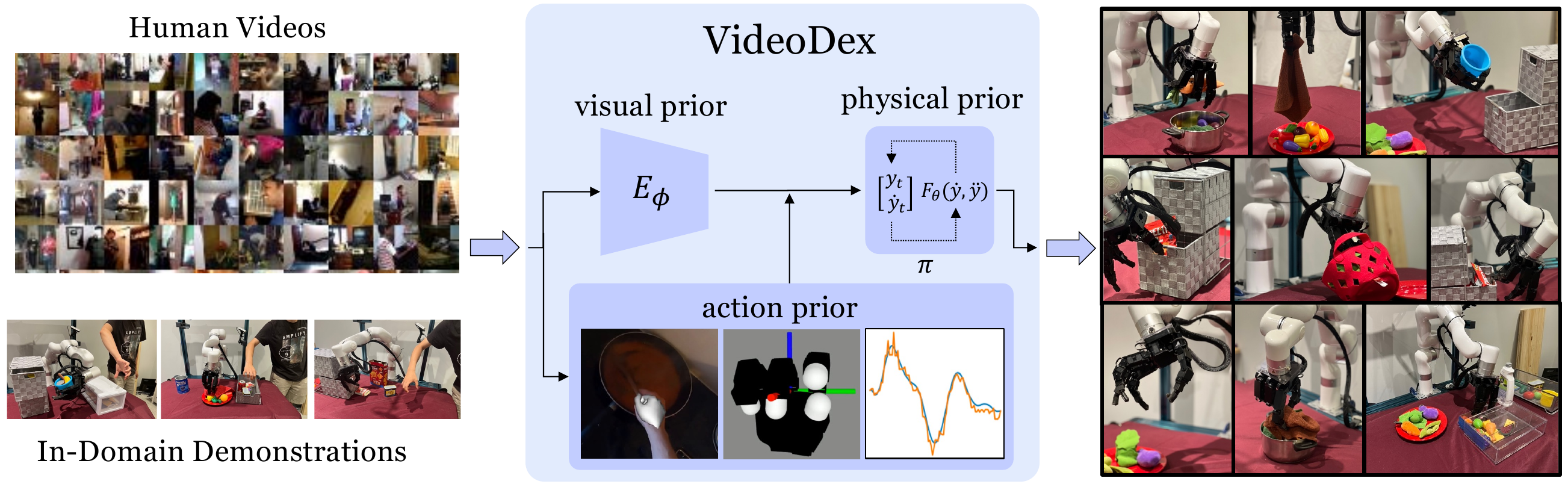}
\caption{\small We re-target human videos as an action prior, use pretrainined embeddings as a visual prior, and use Neural Dynamical Policies (NDPs) \cite{bahl2020neural} as a physical prior to complete many different tasks on a robotic hand.}
 \vspace{-0.15in}
\label{fig:method}
\end{figure}

Fortunately, there is plenty of real-world human interaction videos on the internet. This data can potentially help bootstrap robot learning by side-stepping the data collection-training loop. This insight of leveraging human videos to aid robotics is not new and has seen immense attention from the community at large~\citep{ego4d,EPICKITCHENS,SomethingSomething_ICCV}. However, most of the prior work tends to use human data as a mechanism for pretraining just the visual representation~\citep{pinto2016curious,Sermanet2017TCN,r3m,mvp,nair2018visual}, much like how deep learning has been used as a pretraining tool in related areas of computer vision~\citep{He_2017_ICCV,pmlr-v119-chen20j} and natural language processing~\citep{brown2020language, devlin2018bert}. Although pretraining visual representations can aid in efficiency, we believe that a large part of the inefficiency stems from very large action spaces. For continuous control, learning this is exponential in the number of actions and timesteps, and even more difficult for high degree-of-freedom robots (shown in Figure~\ref{fig:method}).  Dexterous hands are one such class of high degree of freedom robots that have the possibility to provide great contact for the grasping and manipulation of different objects. Their similarity to human hands makes learning from human video advantageous.

In this work, we study how to go beyond using internet human videos merely as a source of visual pretraining (i.e. \textbf{visual priors}), and leverage the information of how humans move their limbs to guide train robots on how they should move (i.e. \textbf{action priors}). However, guiding robot motions using human videos requires understanding the scene in 3D, figuring out human intent, and transferring from human to robot embodiment. First, 3D human estimation works decently well in general human videos which we can leverage to gather 3D understanding.  Second, there have been large-scale datasets that break down the human intent via crowdsourcing labels~\citep{EPICKITCHENS,ego4d}. Finally, to handle the embodiment transfer, we use human hand to robot hand retargeting as an energy function to pretrain the robot action policy. Our key insight is to combine these visual and action priors from human videos with a prior on how robot should move in the world~\citep{bahl2020neural,bahl2021hndp} (i.e., \textbf{physical prior}, using a second order dynamical system) to obtain dexterous robot policies that can act in the real world. We call this approach, \ours. To enhance real-world performance, we mix the experience obtained from massive internet data with a few in-domain demonstrations. 

In summary, \ours is a robot learning algorithm that incorporates visual, action, and physical priors into a single open-loop policy by learning from passive videos contained in human activity datasets from the internet.  \ours then only needs to adapt to real world tasks using a few in-domain examples. We find that \ours outperforms many state-of-the-art robot learning methods on seven different real-world manipulation tasks on a high DOF multi-fingered robotic arm-hand system as well as on a 1-DOf gripper robotic arm system.

\vspace{-0.08in}
\section{Related Work}
\label{sec:relatedwork}

\vspace{-0.1in}
\paragraph{Learning for Dexterity} 
Reinforcement learning (RL) with an engineered reward function can show dexterous simulation results \citep{kalashnikov2018qt, levineFDA15} but requires lots of data, especially in high DOF dexterous manipulation. This requires simulators \citep{todorov12mujoco, makoviychuk2021isaac}, which cannot model physics properly, making real-world transfer difficult.  Behavior cloning is an approach \citep{alvinn, bojarski2016} that can work safely. DIME \citep{dime2022} involves using nearest neighbor matching of image representations with demonstrations to determine actions. \citet{qin2022} teleoperates and learns policies in simulation, followed by Sim2Real transfer. DexMV\citep{qin2021dexmv} uses collected human hand videos for robot hand imitation learning.  DexVIP \citep{dexvip} learns hand-object affordances and priors for RL initialization using curated video datasets.

\vspace{-0.1in}
\paragraph{Learning from Videos and Large-Scale Datasets} There are many curated datasets from internet human videos, for example, FreiHand \citep{Freihand2019} for hand poses, 100 Days of Hands \citep{100doh} for hand-object interactions, Something-Something \citep{SomethingSomething_ICCV} for semantically similar interactions, Human3.6M \citep{ionescu2013human3} and the CMU Mocap Database \citep{cmu_mocap} for Human pose estimation. Epic Kitchens \citep{EPICKITCHENS}, ActivityNet datasets \citep{caba2015activitynet}, or YouCook~\citep{youcook} are action-driven datasets we focus on for dexterous manipulation.

\vspace{-0.1in}
\paragraph{Learning Action from Videos} Detecting humans, estimating poses of different body parts, or understanding the dynamics and interactions related to human motion is a commonly studied problem. One can model human hands using the MANO \citep{MANO:SIGGRAPHASIA:2017} model and the human body using SMPL, SMPL-X \citep{loper2015smpl, SMPL-X:2019} models. There are many efforts in human pose estimation such as \citep{wang2020rgb2hands, hmr, FrankMocap_2021_ICCV}.  We focus on FrankMocap \citep{FrankMocap_2021_ICCV} for our project as it is robust for online videos. Traditionally, teleoperation approaches have employed hand markers with gloves for motion capture \citep{han2018online} or VR settings \citep{MuJoCo_HAPTIX}. Without gloves, Li et. al. \citep{li2019vision} used depth images and a paired human-robot dataset for teleoperation, and Handa et. al. \citep{dex} designed a system that mimics the functional intent of the human operator to perform object manipulation tasks.

\paragraph{Robot Learning by Watching Humans} Recent works have leveraged human datasets to learn cost functions \citep{shao2021concept2robot, chen2021dvd, bahl2022human}, learn action correspondences \cite{schmeckpeper2020reinforcement} both in a paired \citep{sharma2019third} and unpaired manner \citep{smith2019avid}. This data can also be used to extract explicit actions by leveraging structure in the collection (such as reacher-grabber tools \citep{young2020visual}) or prediction of future hand and object locations \citep{lee2017learning}, as well as keypoint detectors \citep{das2020model}. This can also be used to build representations for robot learning \citep{r3m, pari2021surprising}. R3M \citep{r3m} trains on the Ego4D \cite{ego4d} dataset using a temporal alignment loss between language labels and video frames. We build on top of previous efforts in this area, where we combine visual representations trained on human activity data, with \textit{action} driven representations.

\begin{figure}[t]
 \centering
 \includegraphics[width=\linewidth]{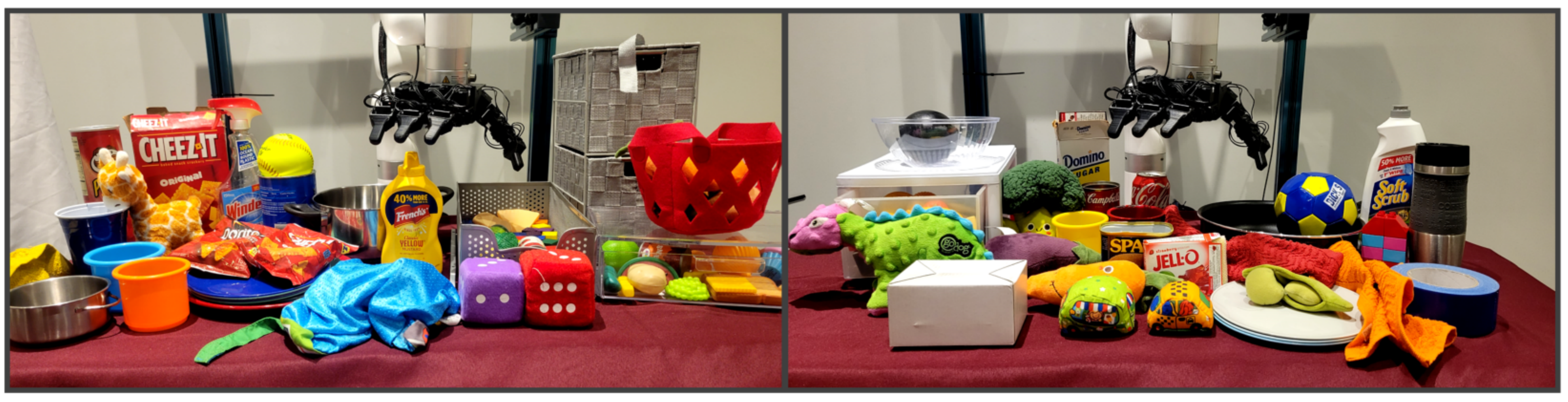}
 \vspace{-0.1in}
\caption{\small The collection of train objects (left) and test objects (right) used for experimentation.}
 \vspace{-0.1in}
\label{fig:objects}
\end{figure}

\vspace{-0.08in}
\section{Background}
\vspace{-0.1in}
\label{section:background}
\subsection{Neural Dynamic Policies}
\vspace{-0.08in}
Neural Dynamic Policies (NDPs) \citep{bahl2020neural, bahl2021hndp, dasari2021rb2}, produce smooth and safe open-loop trajectories.  When using them as a network backbone, they can be rolled out to trajectories of arbitrary lengths which enables the use of varying-length human videos.
NDPs can be described with the Dynamic Movement Primitive equation \citep{isprt2012dmp, prada2013dmp, schaal2006dynamic, pastor2009motorskills}: 
\vspace{-0.02in}
\begin{equation}
    \ddot{y} = \alpha(\beta(g - y) - \dot{y}) + f_w(x, g),
    \label{eq:dmp}
\end{equation}
where $y$ is the coordinate frame of the robot, $g$ is the desired goal in the given coordinate frame, $f_w$ is a radial basis forcing function, $x$ is a time variable, and $\alpha, \beta$ are global constants. NDPs use the robot state, scene, and a NN to output the goal $g$ and shape parameters $w$ of the forcing function $f_w$. 
\vspace{-0.12in}
\subsection{Learning from Watching Humans}
\vspace{-0.07in}
Recently, \citet{robo-telekinesis} introduced Robotic Telekinesis, a pipeline that teleoperates the Allegro Hand \cite{allegro} using a single RGB camera. Leveraging work in monocular human hand and body pose estimation \citep{FrankMocap_2021_ICCV}, hand and body modeling \citep{MANO:SIGGRAPHASIA:2017, loper2015smpl, SMPL-X:2019}, and human internet data, Robotic Telekinesis real-time re-targets the human hand and body to the robot hand and arm. Due to its efficiency and ease of use, we leverage \citet{robo-telekinesis}'s approach for demonstration collection. 

We borrow the human hand to robot hand retargeting method from Robotic Telekinesis~\citep{robo-telekinesis} that manually defines key vectors $v_i^h$ and  $v_i^r$ between palms and fingertips on both the human and robot hand. They build an energy function $E_\pi$ which minimizes the distance between human hand poses  $(\beta, \theta)$ and robot hand poses $q$. $c_i$ is a scale parameter.  Therefore, the energy function is defined as: 
\begin{equation}
    E_{\pi}( \ (\beta_{h}, \theta_{h}), \  q \ ) = \sum_{i=1}^{10} || v_i^h- (c_i \cdot v_i^r) ||_2 ^2
    \label{eq:energy}
\end{equation}
\citet{robo-telekinesis} train an MLP $H_R(.)$ to implicitly minimize this energy function in \ref{eq:energy}, conditioned on knowing human poses  $(\beta, \theta)$. For more details, we refer the readers to \citet{robo-telekinesis}.

\section{Learning Dexterity from Human Videos}

We learn general-purpose manipulation by utilizing large-scale human hand action data as prior robot experience.  We leverage not only visual priors of the scene's appearance but also leverage important aspects of the human hand's motion, intent, and interaction.  To do this, we \textit{re-target} the human video data to trajectories from the robot's embodiment and point of view.  By pretraining policies with these human hand trajectories, we learn \textit{action} priors on how the robot should behave. However, it's notoriously difficult to leverage these noisy human video detections.  Therefore, we must also employ a policy with \textit{physical} priors to learn smooth and robust policies that do not overfit to noise.  We explain insights and our method used to leverage \textit{action} priors in the sections below. 

\vspace{-0.09in}
\subsection{Visual Priors from Human Activity Data}
\vspace{-0.07in}
Many previous works \citep{r3m, mvp, nair2018visual} have tackled visual priors and representations for robot learning.  These networks often encode some form of semantic visual priors into the pretrained network from human video internet datasets.  We use the encoder from \citet{r3m} as a useful visual initialization for our policy.  \citet{r3m} is trained on a visual-language alignment as well as a temporal consistency loss.  Our network takes human video frames and processes them using the publicly released ResNet18 \citep{resnet} encoder, $E_\phi$ from R3M \citep{r3m}.  The output of this network is our visual representation for learning. 
\vspace{-0.07in}
\subsection{Action Priors from Human Activity Data}
\vspace{-0.07in}
\label{action_priors}
While visual pretraining aids in semantic understanding, human data contains a lot more information about how to interact with the world.  \ours uses action information to pretrain an action prior, a network initialization that encodes information about the typical actions for a particular task. 

However, training robot policies on human actions are difficult, as there is a large embodiment gap between humans and robots as described in \citet{dex} and \citet{robo-telekinesis} Thus, we must re-target the motion of the human to the robot embodiment to use it in training.  This problem is solved using three main components. First, we detect human hands in videos. Second, we project hand poses $H$ to robot finger joints $H_r$. Finally, we convert human wrist pose $P$ to robot arm pose $P_r$. $H_r$ and $P_r$ define the trajectory of the human in the robot's frame, from which we can extract actions to pretrain our policy network with the action prior. See Figure~\ref{fig:pipeline} for a summary of the stages.

\label{sec:method}
\vspace{-0.07in}
\begin{figure}[t!]
\centering
\includegraphics[width=\linewidth]{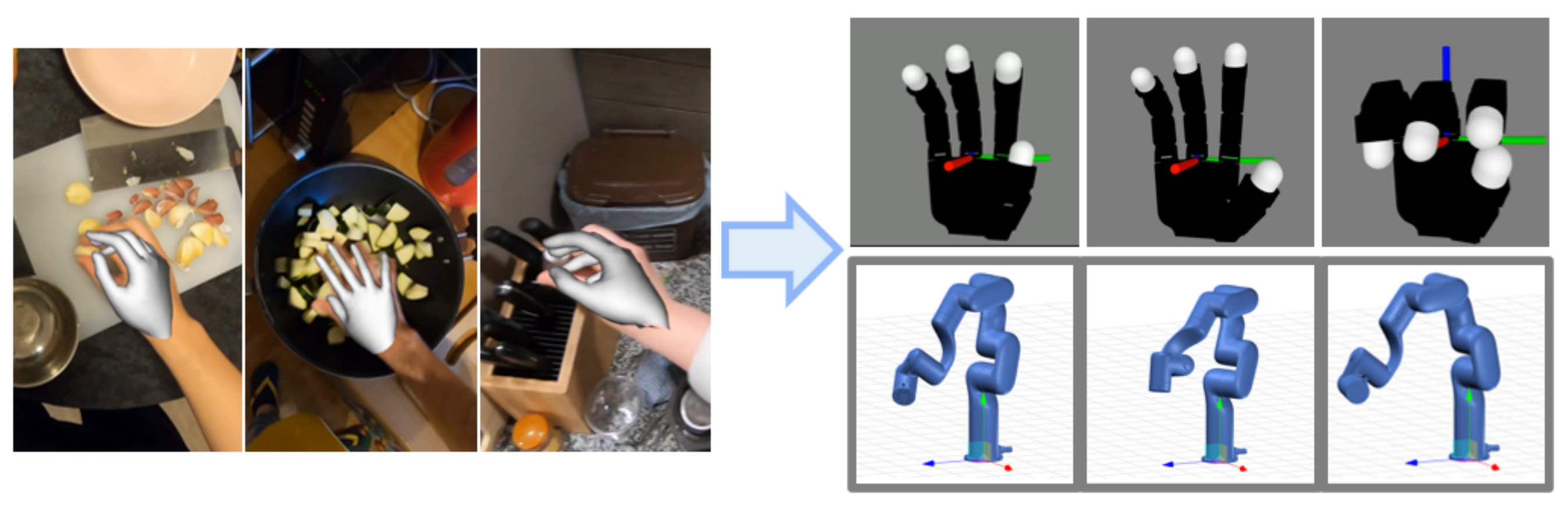}
\vspace{-0.3in}
\caption{\small To use internet videos as pseudo-robot experience, we re-target human hand detections from the 3D MANO model \citep{MANO:SIGGRAPHASIA:2017} to 16 DoF robotic hand (LEAP) embodiment and we retarget the wrist from the moving camera to the xArm6 \cite{xarm} embodiment.  Videos at~\url{https://video-dex.github.io}}
\vspace{-0.1in}
\label{fig:setup}
\end{figure}
\vspace{-0.06in}

\paragraph{Action and Hand Detections} First, we must detect the right actions the human is completing.  To expedite development, we use the action annotations from the EpicKitchens dataset \cite{EPICKITCHENS} but an action detection network such as \cite{2020mmaction2} can be used.  Now, we must detect the hand.  \ours first computes a crop $c$ around the operator's hand using OpenPose \citep{cao2019openpose} and the result is passed to FrankMocap \citep{FrankMocap_2021_ICCV} to obtain hand shape ($\beta$) and pose parameters ($\theta$) of the 3D MANO model \citep{MANO:SIGGRAPHASIA:2017}.  These parameters are passed through a low pass filter and subsequently used in re-targeting to the robot.
\vspace{-0.1in}\paragraph{Re-targeting Wrist Pose}  In this section, we show how to compute the transformation that describes the wrist pose in the robot frame denoted as $M^{Wrist}_{Robot}$.   First, to calculate $M^{Wrist}_{C_t}$, where $C_t$ is the camera frame at timestep $t$ we leverage the Perspective-n-point algorithm \citep{Pnp}. This takes 2D keypoint outputs $(u_i, v_i)$ by the hand detection model and 3D keypoints from the hand model $(x_i, y_i, z_i)$ and computes $M^{Wrist}_{C_t}$. To accurately obtain camera intrinsics for PnP, COLMAP is used \citep{schoenberger2016mvs}.

In human egocentric video datasets, the position of the camera is not fixed and we must compensate for this movement. Specifically, we compute the transformation between the camera pose in the first frame $C_1$ and all other frames in the trajectory, $C_t$. We call this transform $M^{C_t}_{C_1}$. To estimate this, we run monocular SLAM, specifically ORBSLAM3 \citep{ORBSLAM3TRO}. 

Computing wrist poses in the first camera coordinate frame is important but this is still not in the robot frame because the robot is always upright. To be able to transform the human trajectory in the robot's frame, we must find the vector that is parallel to gravity in the camera's frame, $\alpha_p$. Thus recover object segmentations for surfaces that are parallel to the floor such as tables, floors, counters, and similar synonyms using a state-of-the-art object detector (Detic \cite{zhou2022detecting}). Then an estimated depth map from RGB frames only using Adabins \cite{bhat2021adabins} is computed. This way, the method does not rely on the long-term contiguity of a video like most SLAM approaches. We then use depth map portions that correspond to the relevant objects and calculate a surface normal vector. We estimate $\alpha_p$ using this normal vector and the following equations: 
\vspace{-0.10in}
\begin{equation}
\label{eq:pitch}
\text{pitch} = \tan^{-1}(x_{Acc}/ \sqrt{y_{Acc}^2 + z_{Acc}^2})
\end{equation}
\vspace{-0.15in}
\begin{equation}
\label{eq:roll}
\text{roll} = \tan^{-1}(y_{Acc}/ \sqrt{x_{Acc}^2 + z_{Acc}^2})
\end{equation}
\vspace{-0.15in}

Detailed ablations on the parameterization of the initial pitch of the predicted trajectory ( $\alpha$) are provided in Section~\ref{sec:results}. In SLAM, we also remove the dependency on gyroscope data by assuming that the scaling factor is 1.0.  This is acceptable because the trajectory is rescaled to the robot frame later.   Therefore, this wrist re-targeting approach uses only 2D images from human videos.

Since the robot has workspace limits, and we would also like to center the starting pose of the robot, we heuristically compute $T^{World}_{Robot}$ which rescales and rotates the human trajectory in the world frame $\tau_W^\text{wrist}$ into the robot trajectory $\tau_R^\text{wrist}$.  The final function to obtain $M^{Wrist}_{Robot}$ can be described as: 
\begin{equation}
\label{eq:wrist-to-robot}
    M^{Wrist}_{Robot}  = T^{World}_{Robot} \cdot M^{C_1}_{World} \cdot M^{C_t}_{C_1} \cdot M^{wrist}_{C_t}
\end{equation}

\begin{wrapfigure}{l}{0.45\textwidth}
\vspace{-0.3in}
\begin{minipage}{\linewidth}
\begin{algorithm}[H]
\caption{\small Procedure for \ours}\label{alg:cap}
\begin{algorithmic}
\small
\REQUIRE Human videos $V_{1:K}^H$ (length $T$), policy $\pi_\theta$, demonstrations $\mathcal{D}_{1:N}$. Human detection $f_\text{human}$ \citep{FrankMocap_2021_ICCV}. 
\FOR{$k = 1...K$}
    \FOR{$t = 1...T$}
        \STATE Pose parameters $\theta_t, \beta_t$ = $f_\text{human}(I_t)$  
        \STATE Get wrist pose $w_t$ from ~\ref{eq:pitch}, ~\ref{eq:roll} and ~\ref{eq:wrist-to-robot}, 
        \STATE Hand pose $h_t = H(\theta_t, \beta_t)$
    \ENDFOR
    \STATE Store all $h_t$, $w_t$ into robot trajectory $\tau_R^k$
    \STATE $\hat{\tau}_R^k = \pi_\theta(I_1^k, h_1^k, w_1^k)$ 
    \STATE Optimize $\mathcal{L}_\theta = ||\tau_R^k - \hat{\tau}_R^k||_1$ 
\ENDFOR
\STATE Store policy weights $\theta_h$ to initialize $\pi_\theta$
\WHILE{not converged}
\FOR{$n = 1...N$}
    \STATE $\tau_n$, $I_{1:T}^n$ = $\mathcal{D}_n$ 
    \STATE $\hat{\tau}_n = \pi_\theta(I_1^n, h_1^n, w_1^n)$ \
    \STATE Optimize $\mathcal{L}_\theta = ||\tau_n - \hat{\tau}_n||_1$ 
\ENDFOR
\ENDWHILE
\end{algorithmic}
\label{algo}
\end{algorithm}
\end{minipage}
\vspace{-0.35in}
\end{wrapfigure}

\vspace{-0.05in}
\paragraph{Re-targeting Hand Pose} Human hands are also in a different \textit{embodiment} compared to that of robot hands, like our 16 DOF LEAP Hand.  Similarly, to \citet{robo-telekinesis}, we use $H(.)$ to map hand poses to robot hand poses. Given human detected pose $x_h$, we obtain $x_r = H(x_h)$ using a similar re-targeting network to \citet{robo-telekinesis}, and get human hand trajectories: $\tau_R^\text{hand}$ in the robot's embodiment. We use $\tau_R$ to denote the combined hand and wrist trajectories: $\tau_R^\text{hand}$, $\tau_R^\text{wrist}$. See Figure~\ref{fig:setup} for a  visualization.

\begin{figure}[t]
 \centering
 \includegraphics[width=\linewidth]{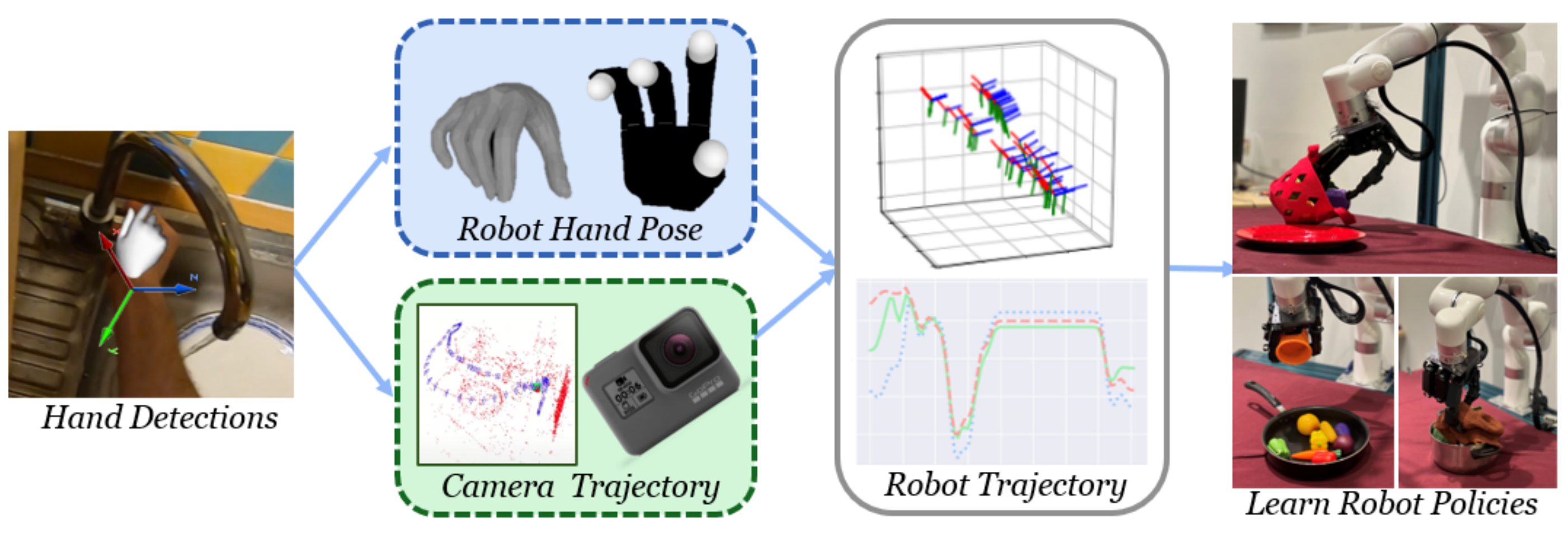}
\caption{\small To use human videos as an action prior for training policies, we re-target them to the robot embodiment. The detected human fingers are converted to the robot fingers using a learned energy function. The wrist is re-targeted using the detections and camera trajectory and transformed to the robot arm.}
 \vspace{-0.15in}
\label{fig:pipeline}
\end{figure}

\vspace{-0.06in}
\subsection{Learning with Human Videos}
\vspace{-0.06in}
We must design an open-loop policy $\pi$ that learns first from the re-targeted human trajectories (the action prior) and then from real robot trajectories collected in teleoperation. Naively, training a neural network policy on $\tau_R$ will lead to overfitting to noisy hand detections. To circumvent this, we first use visual priors from the visual ResNet-based \citep{resnet} encoder provided by \citet{r3m}, $E_\phi$. Then, we introduce a \textit{physical prior} to the network, the physically-inspired Neural Dynamic Policies \citep{bahl2020neural, bahl2021hndp}.

We construct $\pi$ with the following setup. We first process the first scene image $I$ with the visual encoder $E_\phi$. Then the extracted features $E_\phi(I)$ are used to condition an NDP for the wrist and hand separately, $f_{\text{wrist}}$ and $f_{\text{hand}}$. Concretely, each NDP operates by processing the input features with a small MLP which outputs $w, g$ which are the trajectory shape and goal parameters. The forward integrator of the NDP outputs an open-loop trajectory for the hand and the wrist, $\hat{\tau}_R$. We use the following loss function: $$ \mathcal{L} = \sum_k \text{Loss}_{L1}(\tau_R - [f_{\text{hand}}(E_\phi(I_k)), f_{\text{wrist}}(E_\phi(I_k))]) $$ 
\vspace{-0.2in}

\paragraph{Training Methodology:} 
We use between 500-3000 video clips of humans completing the same task category as the robot will from the Epic Kitchens dataset \citep{EPICKITCHENS}. For example, in pick, there are close to 3000 video clips of humans picking items.  These are retargeted to the robot domain and used to pretrain the network with the human action prior of the pick task.  Then, the final policy $\pi$ is trained on a few teleoperated demonstrations of pick on the real robot.  The full training takes about 10 hours on a single 2080Ti GPU.  More training details can be found in the appendix and in Algorithm \ref{algo}.  Our network consists of the R3M \citep{r3m} initialized ResNet-18 \citep{resnet}.  We process these features with a 3-layer MLP with a hidden layer size of 512, which are then processed by 2 NDP \citep{bahl2020neural} networks.

\begin{figure}[t]
\centering
\includegraphics[width=\linewidth]{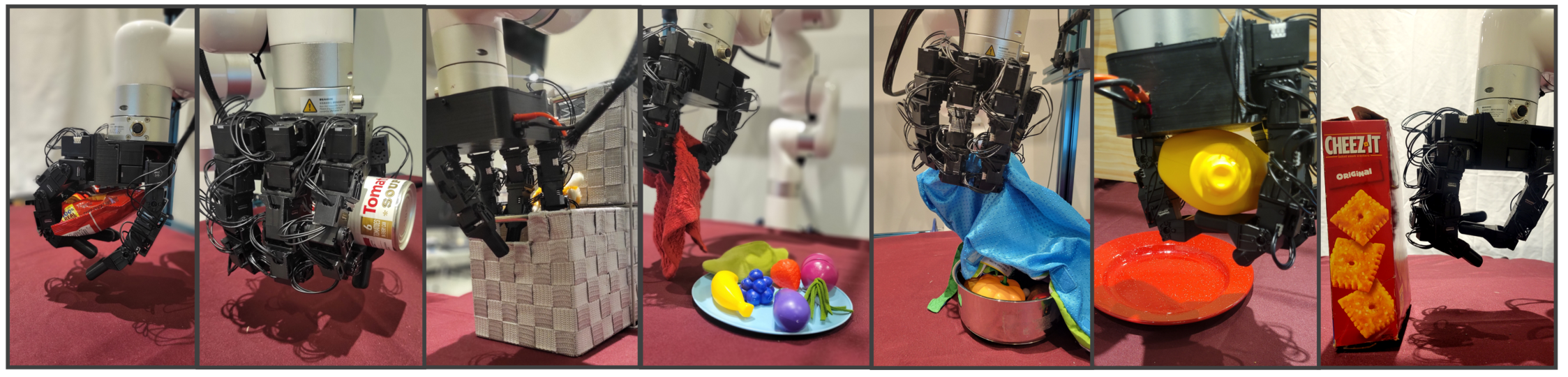}
\vspace{-0.2in}
  \caption{\small Tasks used in experiments. From left to right: pick, rotate, open, cover, uncover, place and push.
  See \url{https://video-dex.github.io} for videos of these tasks.}
 \label{fig:tasks}
 \vspace{-0.15in}
\end{figure}
\vspace{-0.09in}
\section{Experimental Setup}
\vspace{-0.09in}
\label{sec:experiments}
We perform thorough real world experiments on manipulation tasks, specifically many tasks that require dexterity. See \href{https://video-dex.github.io}{our webpage} for result videos.  We aim to answer the following questions. (1) Is \ours able to perform general purpose open-loop manipulation? (2) How much does the action prior of \ours help?  (3) How much does the physical prior of the NDPs in \ours help? (4)  What important design choices are there (visual priors, physical priors, or training setup)? 

\vspace{-0.07in}
\paragraph{Task Setup} We pretrain action priors on retargeted Epic Kitchens data for seven robot tasks. Then, we collect about 120-175 demonstrations for each of these tasks on our setup to train the policy.  In \texttt{pick}, the goal is to pickup an object. In \texttt{rotate}, the agent grasps and rotates the object in place. In \texttt{cover} and \texttt{uncover}, the goal is to cover or uncover a pan/plate with a soft cloth object. \texttt{Push} involves flicking/poking an object with the fingers. In \texttt{place}, the robot has to pick up an object and place it into a plate, pan or pot. In \texttt{open} we open three different drawers. Our testing procedure consists of unseen locations and objects.   Details on the tasks and objects are in the supplemental.

While robot hands can provide great dexterity, we also investigate whether 2-finger grippers can benefit from action priors.    The internet data is converted to where the closed human hand is a closed 2-finger gripper, and the open human hand is an open 2-finger gripper.   We collect separate demonstrations on the real-robot using the 2-finger gripper from xArm \cite{xarm}.  Separate action priors are trained for the 16 DoF LEAP Hand and the 2-finger gripper.  
\vspace{-0.07in}
\section{Results} 
\label{sec:results}
\begin{wrapfigure}{h}{0.58\textwidth}
\vspace{-0.4in}
\centering
\includegraphics[width=\linewidth]{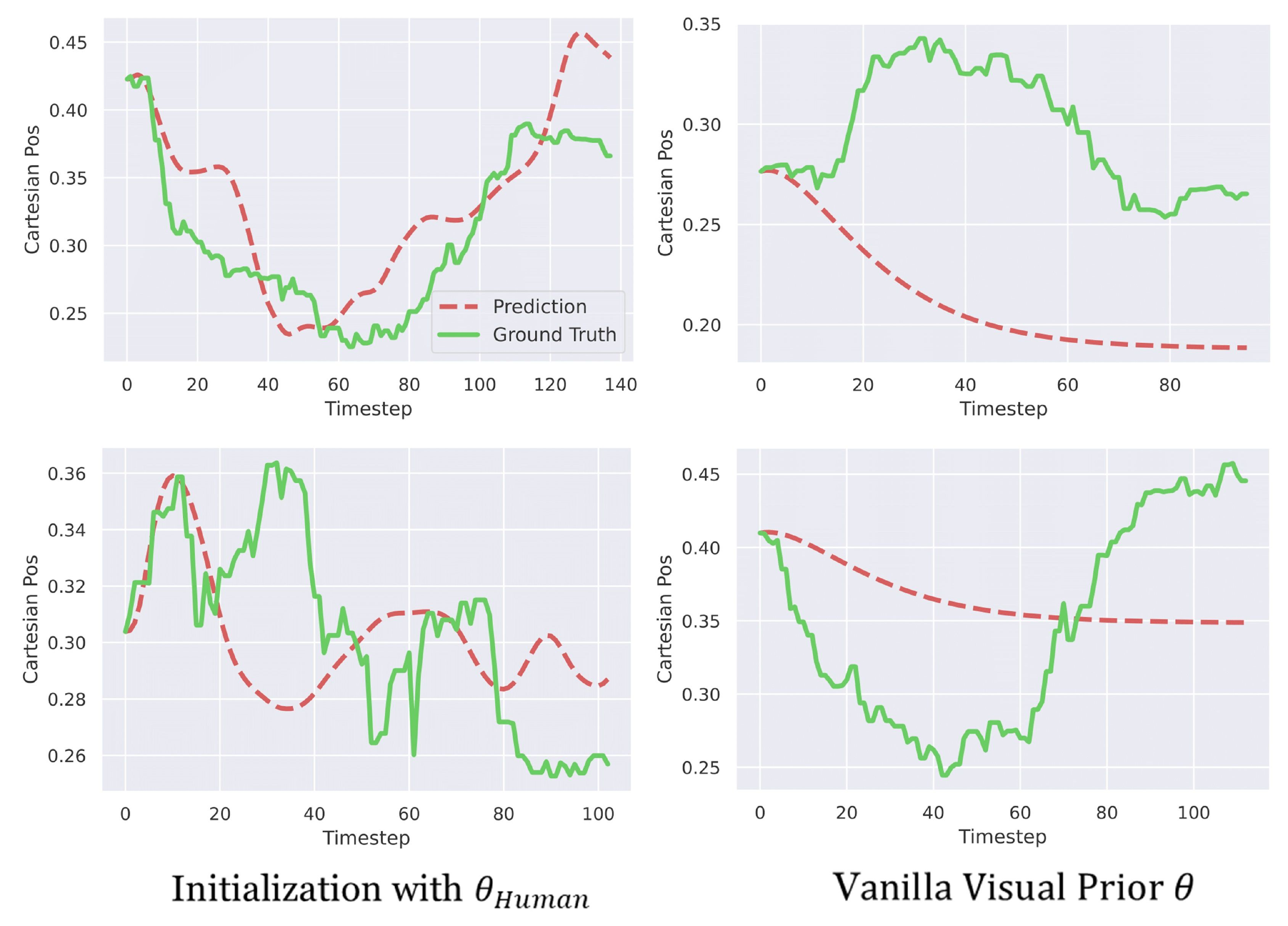}
\vspace{-0.1in}
\caption{\small Networks initialized using action priors on human data without further training are closer to ground truth robot trajectories than networks only initialized using visual priors.}
\label{fig:initialization}
\vspace{-0.1in}
\end{wrapfigure}

First, we evaluate the need for initialization with the action priors obtained from the human internet videos. $\theta_h$ The baseline without internet pre-training is called \texttt{BC-NDP}. It uses the same physical prior and visual network initialization, without the initialization from $\theta_h$.  We also compare the effect of the action prior on 2-finger gripper policies. Second, we compare against two standard open-loop behavior cloning approaches introduced in recent benchmarks \citep{dasari2021rb2}. \texttt{BC-open}  uses a 2 layer MLP instead of the NDP network. \texttt{BC-RNN}, uses an RNN to pre-process the visual features and then a two-stream, 2 layer MLP for wrist and hand trajectories.  We try an offline RL ablation \texttt{CQL} \citep{kumar2020conservative}, where we use the demonstrations as a sparse reward. We train a behavior cloning policy with the action prior from human videos without the physical prior of the NDP. We call this \texttt{\ours-BC-Open}. We ablate the type of visual representation and prior use by trying an initialization using the VGG16 network \citep{simonyan2014very} (\texttt{\ours-VGG}) and the MVP network \citep{mvp} \citep{MAE} (\texttt{\ours-MVP}) based representation trained for robot learning. We ablate the need for a two stream policy, instead training a single NDP for both hand and wrist. (\texttt{\ours-Single}) To see if \ours works with fewer demonstrations (around 50 demonstrations, 5-7 per variant only), we train a policy called \texttt{\ours-Constrained}.

\begin{table}[t]
\centering
\resizebox{\linewidth}{!}{%
\begin{tabular}{lcccccccccccccc}
\toprule
& \multicolumn{2}{c}{Pick} & \multicolumn{2}{c}{Rotate} & \multicolumn{2}{c}{Open} & \multicolumn{2}{c}{Cover}  & \multicolumn{2}{c}{Uncover}  & \multicolumn{2}{c}{Place}  & \multicolumn{2}{c}{Push} \\ 
& train & test & train & test & train & test & train & test & train & test  & train & test  & train & test \\
\midrule
\texttt{BC-NDP} \citep{bahl2021hndp} & 0.64 & 0.38 & \textbf{0.94} & 0.56 & \textbf{0.90} & 0.60 & \textbf{0.78} & 0.58 & 0.88 & 0.82 & 0.70 & 0.35 & 1.00 & 0.71 \\
\texttt{BC-Open}\citep{dasari2021rb2}& 0.50 & 0.44 & 0.72 & 0.38 & 0.80 & 0.40 & 0.44 & 0.58 & \textbf{1.00} & \textbf{0.91} & 0.40 & 0.25 & 1.00 & 0.93 \\
\texttt{BC-RNN} \citep{dasari2021rb2}& 0.56 & 0.31 & 0.78 & 0.50 & \textbf{0.90} & 0.50 & 0.56 & 0.42 & 0.88 & 0.75 & 0.70 & 0.50 & 1.00 & \textbf{1.00} \\
\midrule
\textbf{\texttt{\ours}} & \textbf{0.83} & \textbf{0.77} & 0.85 & \textbf{0.71} & \textbf{0.80} & \textbf{0.80} & \textbf{0.75} & \textbf{0.63}  & \textbf{0.96} & 0.92 & \textbf{0.89} & \textbf{0.80} & 1.00 & \textbf{1.00}\\ 
\bottomrule
\end{tabular}}
\vspace{0.05in}
\caption{\small We present the results of train objects and test objects for Videodex and baselines as described above.}
\vspace{-0.3in}
\label{tab:main}
\end{table}

We analyze the results of our experiments and the guiding questions discussed in Section~\ref{sec:experiments}. We present the results of our findings as a 0-1 success rate in Table~\ref{tab:main} and the result of the ablations we ran on the \texttt{place} task in Table~\ref{tab:abl}.

\vspace{-0.09in}
\paragraph{Effect of Action Priors} We firstly compare \ours against methods that do not employ an action prior trained on human data, as explained in Section~\ref{sec:experiments}. For almost all of the tasks, \ours either outperforms baselines or has a similar performance, especially for held out objects/instances.  We believe that one of the key aspects of \ours generalizing to test objects is the action prior pretraining on human videos. This can be seen in Figure~\ref{fig:initialization}.  Without ever training on the robot demonstrations, the trajectories initialized using the action prior pretrained network $\theta_h$ (left) are much closer to the ground truth trajectories of a network that is initialized using only a visual prior such as the encoder from \citet{r3m} (right). From the results, we see that \texttt{\ours-BC-Open} with action priors (Table~\ref{tab:abl}) outperforms \texttt{BC-Open}.  Having a physical prior added (\texttt{BC-NDP}) tends to help, but it is not the case for every task. We suspect that some tasks require smoother behavior than others. Additionally, in Table~\ref{tab:abl} our offline RL baseline, \texttt{CQL} \citep{kumar2020conservative} does not perform as well as the rest of the approaches, even under-performing the Behavior Cloning setup. Qualitatively, we see a much less smooth and less safe execution with this method, thus we only perform it on one task (\texttt{place}). Note that we use the same visual prior for this as well. 
\begin{wrapfigure}{l}{0.45\textwidth}
\vspace{-0.2in}
\begin{minipage}{0.45\textwidth}
\begin{table}[H]
\resizebox{1.0\linewidth}{!}{%
\begin{tabular}{lccc}
\toprule
 & \textbf{Place} & \textbf{Open} &  \textbf{Pick} \\
\midrule
\texttt{1-DOF BC-Open}\citep{dasari2021rb2}& 0.62& 0.69& 0.71 \\
\midrule
\texttt{1-DOF VideoDex}  & \textbf{0.69} & \textbf{0.82} & \textbf{0.77} \\
\bottomrule
\end{tabular}}
\vspace{0.05in}
\caption{\small We compare how the 1-DOF xArm gripper performs using Videodex. \cite{xarm}  Separate demonstrations were collected using this gripper.}
\label{tab:2finger}
\end{table}
\end{minipage}
\vspace{-0.3in}
\end{wrapfigure}
\vspace{-0.2in}
\paragraph{ Hand vs 2-Finger Gripper}  
We compare whether the action priors from \ours also help in the more general 1-DOF gripper setting.  In Table \ref{tab:2finger}, we find that in the 1-DOF setting, VideoDex still improves performance on these tasks.  This is because the priors from human internet videos still encode typical wrist trajectory behaviors as well as when the gripper should close for each task. 

\begin{wrapfigure}{l}{0.45\textwidth}
\vspace{-0.35in}
\begin{minipage}{0.45\textwidth}
\begin{table}[H]
\resizebox{1.0\linewidth}{!}{%
\begin{tabular}{lccc}
\toprule
 & \textbf{Place} & \textbf{Cover} &  \textbf{Uncover} \\
\midrule
\texttt{\ours-Fixed} & 0.55 & 0.50 & 0.77\\
\texttt{\ours-Random}  & 0.45 & 0.63 & 0.85 \\
\texttt{\ours-IMU}  & 0.70 & \textbf{0.67} & 0.90  \\
\midrule
\texttt{\ours}  & \textbf{0.80} & 0.63 & \textbf{0.92} \\
\bottomrule
\end{tabular}}
\vspace{0.05in}
\caption{\small Ablations that compare the different ways of calculating the initial pitch of the camera with respect to gravity, on test objects. This enables us to transform human trajectories to be upright like the robot is.}
\label{tab:gyro-abl}
\end{table}
\end{minipage}
\vspace{-0.3in}
\end{wrapfigure}

\vspace{-0.09in}
\paragraph{Initial Pose Computation Comparison }
We compare three different ways to estimate $\alpha_p$ or $M^{C_1}_{World}$, the vector that points parallel to gravity.  These methods contrast with \our which uses the surface normal of objects that are typically parallel with the floor to calculate the direction of gravity. \texttt{VideoDex-Fixed}, assumes that $\alpha_p$ is [0,0]. This is reasonable as we are not relying on robots to exactly mimic the human but get a general action prior.  \texttt{VideoDex-Random}, randomizes$\alpha_p$ in the range of 15-45 degrees, which is the typical egocentric camera angle.  \texttt{VideoDex-IMU} uses the internal image stabilization sensor data to estimate the upright vector.  None of these approaches use gyroscope data in SLAM, as we assume that the scaling factor is 1.0. In Table~\ref{tab:gyro-abl}, we present the results of these experiments. The performance degrades when randomizing or setting $M^{C_1}_{World}$ to a fixed value, in all three of the tasks, but it is still comparable to or better than our baselines that do not use any human action data.  A possible explanation for the fact that \texttt{VideoDex-Surface} performed better than our \texttt{\ours-IMU} is that the sensor data may be noisy and estimating surface normals from visual features is more robust. 

\vspace{-0.09in}
\paragraph{Effect of Physical Priors and Architectural Choices} We compare different types of physical priors in Table~\ref{tab:main} and in Table~\ref{tab:abl}. In general (\texttt{BC-NDP}) tends to outperform baselines without a physical prior, except for \texttt{BC-RNN} in a couple of tasks.  \texttt{BC-RNN} performs less aggressive behavior, which allowed it to efficiently grasp more objects. In Table~\ref{tab:abl} it's shown that an important physical prior is to treat the wrist and the hand in a more disentangled manner, as the performance for \texttt{\ours-Single} tends to drop compared to \texttt{BC-NDP} and \texttt{\ours-BC-Open} (Behavior Cloning with our action prior pretraining). The two stream architecture aids in learning, as it allows the policy to disentangle the actions of the wrist and the hand. This is important as the same grasp might be used for picking objects in many different locations, and similarly, it is possible to localize many objects and perform completely different types of interactions.

\begin{wrapfigure}{r}{0.35\textwidth}
\vspace{-0.3in}
\begin{minipage}{0.35\textwidth}
\begin{table}[H]
\resizebox{1.0\linewidth}{!}{%
\begin{tabular}{lcc}
\toprule
 & \textbf{Train} & \textbf{Test} \\
\midrule
\multicolumn{2}{l}{\textit{Baselines}:}\vspace{0.4em}\\
\texttt{BC-NDP} \citep{bahl2021hndp} & 0.70 & 0.35 \\
\texttt{BC-Open} \citep{dasari2021rb2} & 0.40 & 0.25 \\
\texttt{BC-RNN}  \citep{dasari2021rb2} & 0.70 & 0.50 \\
\texttt{CQL}  \citep{kumar2020conservative} & 0.40 & 0.20 \\
\midrule
\multicolumn{2}{l}{\textit{No Physical Prior}:}\vspace{0.4em}\\
\texttt{\ours-BC-Open} & 0.50 & 0.50 \\
\texttt{\ours-Single} & 0.50 & 0.30 \\
\midrule
\multicolumn{2}{l}{\textit{Visual Prior Ablation}:}\vspace{0.4em}\\
\texttt{\ours-VGG} & 0.20 & 0.20\\
\texttt{\ours-MVP} & 0.40 & 0.20 \\
\midrule
\multicolumn{2}{l}{\textit{Constrained Data}:}\vspace{0.4em}\\
\texttt{\ours-Const-5} & 0.80 & 0.60 \\
\texttt{\ours-Const-10} & 0.50 & 0.30 \\
\midrule
\textbf{\texttt{\ours} (ours)}  & \textbf{0.90} & \textbf{0.70}  \\
\bottomrule
\end{tabular}}
\vspace{0.05in}
\caption{\small We present the results of the ablations discussed in Section~\ref{sec:experiments}. These are all performed on the \texttt{place} task.}
\label{tab:abl}
\end{table}
\end{minipage}
\vspace{-0.3in}
\end{wrapfigure}

\vspace{-0.09in}
\paragraph{Generalization with Less Data} We limit \ours to a maximum of 5 and 10 teleoperated demonstrations per variant (we have 12-15 variants in our setup). As shown in Table~\ref{tab:main}, even with 5 instances per variant, we still see a 30\% success rate for unseen objects.  Empirically, the policies generally go to the right area but are not able to grasp objects properly. With less robot experience, \ours outperforms which demonstrates that action priors also boosts sample efficiency.

\vspace{-0.09in}
\paragraph{Effect of Visual Priors} We compared using our approach with MVP (\texttt{\ours-MVP}) \citep{mvp} and VGG (\texttt{\ours-VGG}) \citep{simonyan2014very} and their performance was below \ours using \citet{r3m}. This is likely because both encoders are much larger than the ResNet18 \citep{resnet} we use and require a lot more training time than feasible on human videos. However, \texttt{\ours-MVP} still performs better than \texttt{\ours-VGG}, which indicates that using a visual prior trained on human data does in fact help, as \citet{mvp} trained the representation in self-supervised fashion on videos and use the embeddings to perform robotics tasks in simulation. We see in Table~\ref{tab:main}, that while visual priors are important, action priors are more impactful. 
\vspace{-0.09in}
\paragraph{Choice of Robotic Hand}  
In our experiments, we also tried using the Allegro Hand \cite{allegro}.  We found that the Allegro had higher inaccuracy in control and more hardware failures as compared to LEAP Hand. LEAP Hand outperformed the Allegro Hand $7-12\%$ on average in all experiments, thus we use it for our setup. 

\vspace{-0.12in}
\section{Discussion and Limitations}
\vspace{-0.1in}

Although we see strong results on the held-out objects, \ours has several limitations and scope for future work.  First, we focus on curated human video datasets, such as EpicKitchens \cite{EPICKITCHENS}, but only use these as a convenience to expedite our process. It is possible to filter internet videos of humans according to tasks using action detectors and then processing them with \ours. We also use camera data in \ours but show that with a heuristic driven approach it is possible to obtain similar or better results. Second, we rely on off-the-shelf human hand detection modules that very often have erroneous 6D pose detections, especially when the hand is interacting with objects.  Second, the action priors rely on the arm trajectory as well as the hand trajectory retargeting which must be recomputed for each different set of robot parameters and embodiment.  Finally, our method of behavior cloning in the real world is currently open-loop, so it cannot react to changes in the environment.  This is because closed-loop behavior cloning is difficult to keep safe in the real world. Similarly, when running closed-loop RL it is difficult to guarantee the safety of the system.  We leave this to future work, to train policies that can react to changes in the real world.

\vspace{-0.09in}
\acknowledgments{We thank Aditya Kannan and Shivam Duggal for assisting in robot data collection.  We thank Aravind Sivakumar, Russell Mendonca, Jianren Wang, and Sudeep Dasari for fruitful discussions.  KS is supported by NSF Graduate Research Fellowship under Grant No. DGE2140739.  The work is supported by Samsung GRO Research Award and ONR N00014-22-1-2096.}
\bibliography{main}

\begin{thebibliography}{72}
\providecommand{\natexlab}[1]{#1}
\providecommand{\url}[1]{\texttt{#1}}
\expandafter\ifx\csname urlstyle\endcsname\relax
  \providecommand{\doi}[1]{doi: #1}\else
  \providecommand{\doi}{doi: \begingroup \urlstyle{rm}\Url}\fi

\bibitem[Grauman et~al.(2022)Grauman, Westbury, Byrne, Chavis, Furnari,
  Girdhar, Hamburger, Jiang, Liu, Liu, et~al.]{ego4d}
K.~Grauman, A.~Westbury, E.~Byrne, Z.~Chavis, A.~Furnari, R.~Girdhar,
  J.~Hamburger, H.~Jiang, M.~Liu, X.~Liu, et~al.
\newblock Ego4d: Around the world in 3,000 hours of egocentric video.
\newblock In \emph{Proceedings of the IEEE/CVF Conference on Computer Vision
  and Pattern Recognition}, pages 18995--19012, 2022.

\bibitem[Damen et~al.(2018)Damen, Doughty, Farinella, Fidler, Furnari, Kazakos,
  Moltisanti, Munro, Perrett, Price, and Wray]{EPICKITCHENS}
D.~Damen, H.~Doughty, G.~M. Farinella, S.~Fidler, A.~Furnari, E.~Kazakos,
  D.~Moltisanti, J.~Munro, T.~Perrett, W.~Price, and M.~Wray.
\newblock Scaling egocentric vision: The epic-kitchens dataset.
\newblock In \emph{European Conference on Computer Vision (ECCV)}, 2018.

\bibitem[Goyal et~al.(2017)Goyal, Ebrahimi~Kahou, Michalski, Materzynska,
  Westphal, Kim, Haenel, Fruend, Yianilos, Mueller-Freitag, Hoppe, Thurau, Bax,
  and Memisevic]{SomethingSomething_ICCV}
R.~Goyal, S.~Ebrahimi~Kahou, V.~Michalski, J.~Materzynska, S.~Westphal, H.~Kim,
  V.~Haenel, I.~Fruend, P.~Yianilos, M.~Mueller-Freitag, F.~Hoppe, C.~Thurau,
  I.~Bax, and R.~Memisevic.
\newblock The "something something" video database for learning and evaluating
  visual common sense.
\newblock In \emph{Proceedings of the IEEE International Conference on Computer
  Vision (ICCV)}, Oct 2017.

\bibitem[Pinto et~al.(2016)Pinto, Gandhi, Han, Park, and
  Gupta]{pinto2016curious}
L.~Pinto, D.~Gandhi, Y.~Han, Y.-L. Park, and A.~Gupta.
\newblock The curious robot: Learning visual representations via physical
  interactions.
\newblock In \emph{ECCV}, 2016.

\bibitem[Sermanet et~al.(2018)Sermanet, Lynch, Chebotar, Hsu, Jang, Schaal, and
  Levine]{Sermanet2017TCN}
P.~Sermanet, C.~Lynch, Y.~Chebotar, J.~Hsu, E.~Jang, S.~Schaal, and S.~Levine.
\newblock Time-contrastive networks: Self-supervised learning from video.
\newblock In \emph{ICRA}, 2018.

\bibitem[Nair et~al.(2022)Nair, Rajeswaran, Kumar, Finn, and Gupta]{r3m}
S.~Nair, A.~Rajeswaran, V.~Kumar, C.~Finn, and A.~Gupta.
\newblock R3m: A universal visual representation for robot manipulation.
\newblock \emph{arXiv preprint arXiv:2203.12601}, 2022.

\bibitem[Xiao et~al.(2022)Xiao, Radosavovic, Darrell, and Malik]{mvp}
T.~Xiao, I.~Radosavovic, T.~Darrell, and J.~Malik.
\newblock Masked visual pre-training for motor control.
\newblock \emph{arXiv preprint arXiv:2203.06173}, 2022.

\bibitem[Nair et~al.(2018)Nair, Pong, Dalal, Bahl, Lin, and
  Levine]{nair2018visual}
A.~V. Nair, V.~Pong, M.~Dalal, S.~Bahl, S.~Lin, and S.~Levine.
\newblock Visual reinforcement learning with imagined goals.
\newblock In \emph{NeurIPS}, pages 9191--9200, 2018.

\bibitem[He et~al.(2017)He, Gkioxari, Dollar, and Girshick]{He_2017_ICCV}
K.~He, G.~Gkioxari, P.~Dollar, and R.~Girshick.
\newblock Mask r-cnn.
\newblock In \emph{Proceedings of the IEEE International Conference on Computer
  Vision (ICCV)}, Oct 2017.

\bibitem[Chen et~al.(2020)Chen, Kornblith, Norouzi, and
  Hinton]{pmlr-v119-chen20j}
T.~Chen, S.~Kornblith, M.~Norouzi, and G.~Hinton.
\newblock A simple framework for contrastive learning of visual
  representations.
\newblock In H.~D. III and A.~Singh, editors, \emph{Proceedings of the 37th
  International Conference on Machine Learning}, volume 119 of
  \emph{Proceedings of Machine Learning Research}, pages 1597--1607. PMLR,
  13--18 Jul 2020.
\newblock URL \url{https://proceedings.mlr.press/v119/chen20j.html}.

\bibitem[Brown et~al.(2020)Brown, Mann, Ryder, Subbiah, Kaplan, Dhariwal,
  Neelakantan, Shyam, Sastry, Askell, Agarwal, Herbert-Voss, Krueger, Henighan,
  Child, Ramesh, Ziegler, Wu, Winter, Hesse, Chen, Sigler, Litwin, Gray, Chess,
  Clark, Berner, McCandlish, Radford, Sutskever, and Amodei]{brown2020language}
T.~B. Brown, B.~Mann, N.~Ryder, M.~Subbiah, J.~Kaplan, P.~Dhariwal,
  A.~Neelakantan, P.~Shyam, G.~Sastry, A.~Askell, S.~Agarwal, A.~Herbert-Voss,
  G.~Krueger, T.~Henighan, R.~Child, A.~Ramesh, D.~M. Ziegler, J.~Wu,
  C.~Winter, C.~Hesse, M.~Chen, E.~Sigler, M.~Litwin, S.~Gray, B.~Chess,
  J.~Clark, C.~Berner, S.~McCandlish, A.~Radford, I.~Sutskever, and D.~Amodei.
\newblock Language models are few-shot learners.
\newblock 2020.

\bibitem[Devlin et~al.(2018)Devlin, Chang, Lee, and Toutanova]{devlin2018bert}
J.~Devlin, M.-W. Chang, K.~Lee, and K.~Toutanova.
\newblock Bert: Pre-training of deep bidirectional transformers for language
  understanding.
\newblock \emph{arXiv preprint arXiv:1810.04805}, 2018.

\bibitem[Bahl et~al.(2020)Bahl, Mukadam, Gupta, and Pathak]{bahl2020neural}
S.~Bahl, M.~Mukadam, A.~Gupta, and D.~Pathak.
\newblock Neural dynamic policies for end-to-end sensorimotor learning.
\newblock In \emph{NeurIPS}, 2020.

\bibitem[Bahl et~al.(2021)Bahl, Gupta, and Pathak]{bahl2021hndp}
S.~Bahl, A.~Gupta, and D.~Pathak.
\newblock Hierarchical neural dynamic policies.
\newblock \emph{RSS}, 2021.

\bibitem[Kalashnikov et~al.(2018)Kalashnikov, Irpan, Pastor, Ibarz, Herzog,
  Jang, Quillen, Holly, Kalakrishnan, Vanhoucke, et~al.]{kalashnikov2018qt}
D.~Kalashnikov, A.~Irpan, P.~Pastor, J.~Ibarz, A.~Herzog, E.~Jang, D.~Quillen,
  E.~Holly, M.~Kalakrishnan, V.~Vanhoucke, et~al.
\newblock Qt-opt: Scalable deep reinforcement learning for vision-based robotic
  manipulation.
\newblock \emph{arXiv preprint arXiv:1806.10293}, 2018.

\bibitem[Levine et~al.(2016)Levine, Finn, Darrell, and Abbeel]{levineFDA15}
S.~Levine, C.~Finn, T.~Darrell, and P.~Abbeel.
\newblock End-to-end training of deep visuomotor policies.
\newblock \emph{JMLR}, 2016.

\bibitem[Todorov et~al.(2012)Todorov, Erez, and Tassa]{todorov12mujoco}
E.~Todorov, T.~Erez, and Y.~Tassa.
\newblock {MuJoCo: A physics engine for model-based control}.
\newblock In \emph{IROS}, 2012.

\bibitem[Makoviychuk et~al.(2021)Makoviychuk, Wawrzyniak, Guo, Lu, Storey,
  Macklin, Hoeller, Rudin, Allshire, Handa, et~al.]{makoviychuk2021isaac}
V.~Makoviychuk, L.~Wawrzyniak, Y.~Guo, M.~Lu, K.~Storey, M.~Macklin,
  D.~Hoeller, N.~Rudin, A.~Allshire, A.~Handa, et~al.
\newblock Isaac gym: High performance gpu-based physics simulation for robot
  learning.
\newblock \emph{arXiv preprint arXiv:2108.10470}, 2021.

\bibitem[Pomerleau(1988)]{alvinn}
D.~A. Pomerleau.
\newblock Alvinn: An autonomous land vehicle in a neural network.
\newblock In D.~Touretzky, editor, \emph{Advances in Neural Information
  Processing Systems}, volume~1. Morgan-Kaufmann, 1988.
\newblock URL
  \url{https://proceedings.neurips.cc/paper/1988/file/812b4ba287f5ee0bc9d43bbf5bbe87fb-Paper.pdf}.

\bibitem[Bojarski et~al.(2016)Bojarski, Del~Testa, Dworakowski, Firner, Flepp,
  Goyal, Jackel, Monfort, Muller, Zhang, Zhang, Zhao, and Zieba]{bojarski2016}
M.~Bojarski, D.~Del~Testa, D.~Dworakowski, B.~Firner, B.~Flepp, P.~Goyal, L.~D.
  Jackel, M.~Monfort, U.~Muller, J.~Zhang, X.~Zhang, J.~Zhao, and K.~Zieba.
\newblock End to end learning for self-driving cars, 2016.
\newblock URL \url{https://arxiv.org/abs/1604.07316}.

\bibitem[Arunachalam et~al.(2022)Arunachalam, Silwal, Evans, and
  Pinto]{dime2022}
S.~P. Arunachalam, S.~Silwal, B.~Evans, and L.~Pinto.
\newblock Dexterous imitation made easy: A learning-based framework for
  efficient dexterous manipulation, 2022.
\newblock URL \url{https://arxiv.org/abs/2203.13251}.

\bibitem[Qin et~al.(2022)Qin, Su, and Wang]{qin2022}
Y.~Qin, H.~Su, and X.~Wang.
\newblock From one hand to multiple hands: Imitation learning for dexterous
  manipulation from single-camera teleoperation, 2022.
\newblock URL \url{https://arxiv.org/abs/2204.12490}.

\bibitem[Qin et~al.(2021)Qin, Wu, Liu, Jiang, Yang, Fu, and Wang]{qin2021dexmv}
Y.~Qin, Y.-H. Wu, S.~Liu, H.~Jiang, R.~Yang, Y.~Fu, and X.~Wang.
\newblock Dexmv: Imitation learning for dexterous manipulation from human
  videos.
\newblock \emph{arXiv preprint arXiv:2108.05877}, 2021.

\bibitem[Mandikal and Grauman(2022)]{dexvip}
P.~Mandikal and K.~Grauman.
\newblock Dexvip: Learning dexterous grasping with human hand pose priors from
  video.
\newblock In \emph{Conference on Robot Learning}, pages 651--661. PMLR, 2022.

\bibitem[Zimmermann et~al.(2019)Zimmermann, Ceylan, Yang, Russell, Argus, and
  Brox]{Freihand2019}
C.~Zimmermann, D.~Ceylan, J.~Yang, B.~Russell, M.~Argus, and T.~Brox.
\newblock Freihand: A dataset for markerless capture of hand pose and shape
  from single rgb images.
\newblock In \emph{Proceedings of the IEEE/CVF International Conference on
  Computer Vision}, pages 813--822, 2019.

\bibitem[Shan et~al.(2020)Shan, Geng, Shu, and Fouhey]{100doh}
D.~Shan, J.~Geng, M.~Shu, and D.~F. Fouhey.
\newblock Understanding human hands in contact at internet scale.
\newblock In \emph{Proceedings of the IEEE/CVF Conference on Computer Vision
  and Pattern Recognition}, pages 9869--9878, 2020.

\bibitem[Ionescu et~al.(2013)Ionescu, Papava, Olaru, and
  Sminchisescu]{ionescu2013human3}
C.~Ionescu, D.~Papava, V.~Olaru, and C.~Sminchisescu.
\newblock Human3. 6m: Large scale datasets and predictive methods for 3d human
  sensing in natural environments.
\newblock \emph{IEEE transactions on pattern analysis and machine
  intelligence}, 36\penalty0 (7):\penalty0 1325--1339, 2013.

\bibitem[cmu()]{cmu_mocap}
Cmu graphics lab motion capture database.
\newblock \url{http://mocap.cs.cmu.edu/}.

\bibitem[Fabian Caba~Heilbron and Niebles(2015)]{caba2015activitynet}
B.~G. Fabian Caba~Heilbron, Victor~Escorcia and J.~C. Niebles.
\newblock Activitynet: A large-scale video benchmark for human activity
  understanding.
\newblock In \emph{CVPR}, pages 961--970, 2015.

\bibitem[Das et~al.(2013)Das, Xu, Doell, and Corso]{youcook}
P.~Das, C.~Xu, R.~F. Doell, and J.~J. Corso.
\newblock A thousand frames in just a few words: Lingual description of videos
  through latent topics and sparse object stitching.
\newblock In \emph{Proceedings of the IEEE conference on computer vision and
  pattern recognition}, pages 2634--2641, 2013.

\bibitem[Romero et~al.(2017)Romero, Tzionas, and Black]{MANO:SIGGRAPHASIA:2017}
J.~Romero, D.~Tzionas, and M.~J. Black.
\newblock Embodied hands: Modeling and capturing hands and bodies together.
\newblock \emph{ACM Transactions on Graphics, (Proc. SIGGRAPH Asia)},
  36\penalty0 (6), Nov. 2017.

\bibitem[Loper et~al.(2015)Loper, Mahmood, Romero, Pons-Moll, and
  Black]{loper2015smpl}
M.~Loper, N.~Mahmood, J.~Romero, G.~Pons-Moll, and M.~J. Black.
\newblock Smpl: A skinned multi-person linear model.
\newblock \emph{ACM transactions on graphics (TOG)}, 34\penalty0 (6):\penalty0
  1--16, 2015.

\bibitem[Pavlakos et~al.(2019)Pavlakos, Choutas, Ghorbani, Bolkart, Osman,
  Tzionas, and Black]{SMPL-X:2019}
G.~Pavlakos, V.~Choutas, N.~Ghorbani, T.~Bolkart, A.~A.~A. Osman, D.~Tzionas,
  and M.~J. Black.
\newblock Expressive body capture: {3D} hands, face, and body from a single
  image.
\newblock In \emph{Proceedings IEEE Conf. on Computer Vision and Pattern
  Recognition (CVPR)}, pages 10975--10985, 2019.

\bibitem[Wang et~al.(2020)Wang, Mueller, Bernard, Sorli, Sotnychenko, Qian,
  Otaduy, Casas, and Theobalt]{wang2020rgb2hands}
J.~Wang, F.~Mueller, F.~Bernard, S.~Sorli, O.~Sotnychenko, N.~Qian, M.~A.
  Otaduy, D.~Casas, and C.~Theobalt.
\newblock Rgb2hands: real-time tracking of 3d hand interactions from monocular
  rgb video.
\newblock \emph{ACM Transactions on Graphics (TOG)}, 39\penalty0 (6):\penalty0
  1--16, 2020.

\bibitem[Kanazawa et~al.(2017)Kanazawa, Black, Jacobs, and Malik]{hmr}
A.~Kanazawa, M.~J. Black, D.~W. Jacobs, and J.~Malik.
\newblock End-to-end recovery of human shape and pose.
\newblock \emph{CoRR}, abs/1712.06584, 2017.
\newblock URL \url{http://arxiv.org/abs/1712.06584}.

\bibitem[Rong et~al.(2021)Rong, Shiratori, and Joo]{FrankMocap_2021_ICCV}
Y.~Rong, T.~Shiratori, and H.~Joo.
\newblock Frankmocap: A monocular 3d whole-body pose estimation system via
  regression and integration.
\newblock In \emph{Proceedings of the IEEE/CVF International Conference on
  Computer Vision (ICCV) Workshops}, pages 1749--1759, October 2021.

\bibitem[Han et~al.(2018)Han, Liu, Wang, Ye, Twigg, and Kin]{han2018online}
S.~Han, B.~Liu, R.~Wang, Y.~Ye, C.~D. Twigg, and K.~Kin.
\newblock Online optical marker-based hand tracking with deep labels.
\newblock \emph{ACM Transactions on Graphics (TOG)}, 37\penalty0 (4):\penalty0
  1--10, 2018.

\bibitem[Kumar and Todorov(2015)]{MuJoCo_HAPTIX}
V.~Kumar and E.~Todorov.
\newblock Mujoco haptix: A virtual reality system for hand manipulation.
\newblock In \emph{2015 IEEE-RAS 15th International Conference on Humanoid
  Robots (Humanoids)}, pages 657--663, 2015.
\newblock \doi{10.1109/HUMANOIDS.2015.7363441}.

\bibitem[Li et~al.(2019)Li, Ma, Liang, G{\"o}rner, Ruppel, Fang, Sun, and
  Zhang]{li2019vision}
S.~Li, X.~Ma, H.~Liang, M.~G{\"o}rner, P.~Ruppel, B.~Fang, F.~Sun, and
  J.~Zhang.
\newblock Vision-based teleoperation of shadow dexterous hand using end-to-end
  deep neural network.
\newblock In \emph{2019 International Conference on Robotics and Automation
  (ICRA)}, pages 416--422. IEEE, 2019.

\bibitem[Handa et~al.(2020)Handa, Van~Wyk, Yang, Liang, Chao, Wan, Birchfield,
  Ratliff, and Fox]{dex}
A.~Handa, K.~Van~Wyk, W.~Yang, J.~Liang, Y.-W. Chao, Q.~Wan, S.~Birchfield,
  N.~Ratliff, and D.~Fox.
\newblock Dexpilot: Vision-based teleoperation of dexterous robotic hand-arm
  system.
\newblock In \emph{2020 IEEE International Conference on Robotics and
  Automation (ICRA)}, pages 9164--9170, 2020.
\newblock \doi{10.1109/ICRA40945.2020.9197124}.

\bibitem[Shao et~al.(2021)Shao, Migimatsu, Zhang, Yang, and
  Bohg]{shao2021concept2robot}
L.~Shao, T.~Migimatsu, Q.~Zhang, K.~Yang, and J.~Bohg.
\newblock Concept2robot: Learning manipulation concepts from instructions and
  human demonstrations.
\newblock \emph{The International Journal of Robotics Research}, 40\penalty0
  (12-14), 2021.

\bibitem[Chen et~al.(2021)Chen, Nair, and Finn]{chen2021dvd}
A.~S. Chen, S.~Nair, and C.~Finn.
\newblock Learning generalizable robotic reward functions from" in-the-wild"
  human videos.
\newblock \emph{arXiv preprint arXiv:2103.16817}, 2021.

\bibitem[Bahl et~al.(2022)Bahl, Gupta, and Pathak]{bahl2022human}
S.~Bahl, A.~Gupta, and D.~Pathak.
\newblock Human-to-robot imitation in the wild.
\newblock \emph{RSS}, 2022.

\bibitem[Schmeckpeper et~al.(2020)Schmeckpeper, Rybkin, Daniilidis, Levine, and
  Finn]{schmeckpeper2020reinforcement}
K.~Schmeckpeper, O.~Rybkin, K.~Daniilidis, S.~Levine, and C.~Finn.
\newblock Reinforcement learning with videos: Combining offline observations
  with interaction.
\newblock \emph{arXiv preprint arXiv:2011.06507}, 2020.

\bibitem[Sharma et~al.(2019)Sharma, Pathak, and Gupta]{sharma2019third}
P.~Sharma, D.~Pathak, and A.~Gupta.
\newblock Third-person visual imitation learning via decoupled hierarchical
  controller.
\newblock \emph{arXiv preprint arXiv:1911.09676}, 2019.

\bibitem[Smith et~al.(2020)Smith, Dhawan, Zhang, Abbeel, and
  Levine]{smith2019avid}
L.~Smith, N.~Dhawan, M.~Zhang, P.~Abbeel, and S.~Levine.
\newblock Avid: Learning multi-stage tasks via pixel-level translation of human
  videos.
\newblock In \emph{RSS}, 2020.

\bibitem[Young et~al.(2020)Young, Gandhi, Tulsiani, Gupta, Abbeel, and
  Pinto]{young2020visual}
S.~Young, D.~Gandhi, S.~Tulsiani, A.~Gupta, P.~Abbeel, and L.~Pinto.
\newblock Visual imitation made easy.
\newblock \emph{arXiv preprint arXiv:2008.04899}, 2020.

\bibitem[Lee and Ryoo(2017)]{lee2017learning}
J.~Lee and M.~S. Ryoo.
\newblock Learning robot activities from first-person human videos using
  convolutional future regression.
\newblock In \emph{CVPR Workshops}, pages 1--2, 2017.

\bibitem[Das et~al.(2020)Das, Bechtle, Davchev, Jayaraman, Rai, and
  Meier]{das2020model}
N.~Das, S.~Bechtle, T.~Davchev, D.~Jayaraman, A.~Rai, and F.~Meier.
\newblock Model-based inverse reinforcement learning from visual
  demonstrations.
\newblock \emph{arXiv preprint arXiv:2010.09034}, 2020.

\bibitem[Pari et~al.(2021)Pari, Muhammad, Arunachalam, Pinto,
  et~al.]{pari2021surprising}
J.~Pari, N.~Muhammad, S.~P. Arunachalam, L.~Pinto, et~al.
\newblock The surprising effectiveness of representation learning for visual
  imitation.
\newblock \emph{arXiv preprint arXiv:2112.01511}, 2021.

\bibitem[Dasari et~al.(2021)Dasari, Wang, Hong, Bahl, Lin, Wang, Thankaraj,
  Chahal, Calli, Gupta, et~al.]{dasari2021rb2}
S.~Dasari, J.~Wang, J.~Hong, S.~Bahl, Y.~Lin, A.~S. Wang, A.~Thankaraj, K.~S.
  Chahal, B.~Calli, S.~Gupta, et~al.
\newblock Rb2: Robotic manipulation benchmarking with a twist.
\newblock In \emph{NeurIPS Datasets and Benchmarks Track (Round 2)}, 2021.

\bibitem[Ijspeert et~al.(2013)Ijspeert, Nakanishi, Hoffmann, Pastor, and
  Schaal]{isprt2012dmp}
A.~J. Ijspeert, J.~Nakanishi, H.~Hoffmann, P.~Pastor, and S.~Schaal.
\newblock Dynamical movement primitives: Learning attractor models for motor
  behaviors.
\newblock \emph{Neural Computation}, 2013.

\bibitem[{Prada} et~al.(2013){Prada}, {Remazeilles}, {Koene}, and
  {Endo}]{prada2013dmp}
M.~{Prada}, A.~{Remazeilles}, A.~{Koene}, and S.~{Endo}.
\newblock Dynamic movement primitives for human-robot interaction: Comparison
  with human behavioral observation.
\newblock In \emph{International Conference on Intelligent Robots and Systems},
  2013.

\bibitem[Schaal(2006)]{schaal2006dynamic}
S.~Schaal.
\newblock Dynamic movement primitives-a framework for motor control in humans
  and humanoid robotics.
\newblock In \emph{Adaptive motion of animals and machines}. Springer, 2006.

\bibitem[Pastor et~al.(2009)Pastor, Hoffmann, Asfour, and
  Schaal]{pastor2009motorskills}
P.~Pastor, H.~Hoffmann, T.~Asfour, and S.~Schaal.
\newblock Learning and generalization of motor skills by learning from
  demonstration.
\newblock In \emph{ICRA}, 2009.

\bibitem[Sivakumar et~al.(2022)Sivakumar, Shaw, and Pathak]{robo-telekinesis}
A.~Sivakumar, K.~Shaw, and D.~Pathak.
\newblock Robotic telekinesis: Learning a robotic hand imitator by watching
  humans on youtube, 2022.

\bibitem[all()]{allegro}
Allegro hand.
\newblock \url{https://www.wonikrobotics.com/research-robot-hand}.

\bibitem[He et~al.(2015)He, Zhang, Ren, and Sun]{resnet}
K.~He, X.~Zhang, S.~Ren, and J.~Sun.
\newblock Deep residual learning for image recognition.
\newblock \emph{CoRR}, abs/1512.03385, 2015.
\newblock URL \url{http://arxiv.org/abs/1512.03385}.

\bibitem[xar()]{xarm}
xarm6 by ufactory.
\newblock \url{https://www.ufactory.cc/xarm-collaborative-robot}.

\bibitem[Contributors(2020)]{2020mmaction2}
M.~Contributors.
\newblock Openmmlab's next generation video understanding toolbox and
  benchmark.
\newblock \url{https://github.com/open-mmlab/mmaction2}, 2020.

\bibitem[Cao et~al.(2019)Cao, Hidalgo, Simon, Wei, and Sheikh]{cao2019openpose}
Z.~Cao, G.~Hidalgo, T.~Simon, S.-E. Wei, and Y.~Sheikh.
\newblock Openpose: realtime multi-person 2d pose estimation using part
  affinity fields.
\newblock \emph{IEEE transactions on pattern analysis and machine
  intelligence}, 43\penalty0 (1):\penalty0 172--186, 2019.

\bibitem[Fischler and Bolles(1981)]{Pnp}
M.~A. Fischler and R.~C. Bolles.
\newblock Random sample consensus: A paradigm for model fitting with
  applications to image analysis and automated cartography.
\newblock \emph{Commun. ACM}, 24\penalty0 (6):\penalty0 381–395, jun 1981.
\newblock ISSN 0001-0782.
\newblock \doi{10.1145/358669.358692}.
\newblock URL \url{https://doi.org/10.1145/358669.358692}.

\bibitem[Sch\"{o}nberger et~al.(2016)Sch\"{o}nberger, Zheng, Pollefeys, and
  Frahm]{schoenberger2016mvs}
J.~L. Sch\"{o}nberger, E.~Zheng, M.~Pollefeys, and J.-M. Frahm.
\newblock {Pixelwise View Selection for Unstructured Multi-View Stereo}.
\newblock In \emph{European Conference on Computer Vision (ECCV)}, 2016.

\bibitem[Campos et~al.(2021)Campos, Elvira, Rodr{\'\i}guez, Montiel, and
  Tard{\'o}s]{ORBSLAM3TRO}
C.~Campos, R.~Elvira, J.~J.~G. Rodr{\'\i}guez, J.~M. Montiel, and J.~D.
  Tard{\'o}s.
\newblock Orb-slam3: An accurate open-source library for visual,
  visual--inertial, and multimap slam.
\newblock \emph{IEEE Transactions on Robotics}, 37\penalty0 (6):\penalty0
  1874--1890, 2021.

\bibitem[Zhou et~al.(2022)Zhou, Girdhar, Joulin, Kr{\"a}henb{\"u}hl, and
  Misra]{zhou2022detecting}
X.~Zhou, R.~Girdhar, A.~Joulin, P.~Kr{\"a}henb{\"u}hl, and I.~Misra.
\newblock Detecting twenty-thousand classes using image-level supervision.
\newblock In \emph{ECCV}, 2022.

\bibitem[Bhat et~al.(2021)Bhat, Alhashim, and Wonka]{bhat2021adabins}
S.~F. Bhat, I.~Alhashim, and P.~Wonka.
\newblock Adabins: Depth estimation using adaptive bins.
\newblock In \emph{Proceedings of the IEEE/CVF Conference on Computer Vision
  and Pattern Recognition}, pages 4009--4018, 2021.

\bibitem[Kumar et~al.(2020)Kumar, Zhou, Tucker, and
  Levine]{kumar2020conservative}
A.~Kumar, A.~Zhou, G.~Tucker, and S.~Levine.
\newblock Conservative q-learning for offline reinforcement learning.
\newblock \emph{Advances in Neural Information Processing Systems},
  33:\penalty0 1179--1191, 2020.

\bibitem[Simonyan and Zisserman(2014)]{simonyan2014very}
K.~Simonyan and A.~Zisserman.
\newblock Very deep convolutional networks for large-scale image recognition.
\newblock \emph{arXiv preprint arXiv:1409.1556}, 2014.

\bibitem[He et~al.(2022)He, Chen, Xie, Li, Doll{\'a}r, and Girshick]{MAE}
K.~He, X.~Chen, S.~Xie, Y.~Li, P.~Doll{\'a}r, and R.~Girshick.
\newblock Masked autoencoders are scalable vision learners.
\newblock In \emph{Proceedings of the IEEE/CVF Conference on Computer Vision
  and Pattern Recognition}, pages 16000--16009, 2022.

\bibitem[Deng et~al.(2009)Deng, Dong, Socher, Li, Li, and
  Fei-Fei]{deng2009imagenet}
J.~Deng, W.~Dong, R.~Socher, L.-J. Li, K.~Li, and L.~Fei-Fei.
\newblock Imagenet: A large-scale hierarchical image database.
\newblock In \emph{2009 IEEE conference on computer vision and pattern
  recognition}, pages 248--255. Ieee, 2009.

\bibitem[Handa et~al.(2020)Handa, Van~Wyk, Yang, Liang, Chao, Wan, Birchfield,
  Ratliff, and Fox]{handa2020dexpilot}
A.~Handa, K.~Van~Wyk, W.~Yang, J.~Liang, Y.-W. Chao, Q.~Wan, S.~Birchfield,
  N.~Ratliff, and D.~Fox.
\newblock Dexpilot: Vision-based teleoperation of dexterous robotic hand-arm
  system.
\newblock In \emph{2020 IEEE International Conference on Robotics and
  Automation (ICRA)}, pages 9164--9170. IEEE, 2020.

\bibitem[Takuma~Seno(2021)]{seno2021d3rlpy}
M.~I. Takuma~Seno.
\newblock d3rlpy: An offline deep reinforcement library.
\newblock In \emph{NeurIPS 2021 Offline Reinforcement Learning Workshop},
  December 2021.

\end{thebibliography}

\clearpage
\appendix

\section{Videos}

Task videos, performance videos, data collection and example of internet videos can be found at: \url{https://videodex.github.io}

\section{Additional Ablations}

\textbf{Comparing Effects of Actions, Visual and Physical Priors:} Firstly, we ran an ablation where we pertained a policy on human videos performing the place task and finetune it on the uncover task (using robot data). Similarly, we pretrained a policy on Uncover and finetuned on place. The results are in the below table under \texttt{VideoDex-Transfer}. We see that for both tasks the performance degrades slightly, especially in the place task. We also train by adding noise to the demonstration trajectories, by adding two different levels of Gaussian noise with standard deviation being 0.01 and 0.05, shown as \texttt{VideoDex-Noise-0.01} and \texttt{VideoDex-Noise-0.05}. We find that adding more noise definitely hurts the performance of the method. We also train ResNet18 \cite{resnet} features initialized from ImageNet \cite{deng2009imagenet} training instead of the R3M \cite{r3m} features, and the results in \texttt{VideoDex-ImageNet}. We can see that performance drops off, which indicates that the visual priors are important. Note that all of the reported numbers are on test objects. We present the results in Table~\ref{tab:ndp-abl}.

\begin{table}[b]
\centering
\begin{tabular}{lcc}
\toprule
 Transforms & Description & Method \\
\midrule
{\large$M^{Wrist}_{C_t}$}& Wrist in each Camera & FrankMocap + PnP \\ [1.5ex]
{\large$M^{C_t}_{C_1}$} & Track Moving Camera & IMU/ORBSLAM \\ [1.5ex] 
{\large$M^{C_1}_{World}$} & Make Camera parallel to Ground & IMU/Stabilization Sensor\\ [1.5ex] 
{\large$T^{World}_{Robot}$} & Rescale and Reorient for Robot & Heuristic\\ [1.5ex] 
\midrule
{\large$M^{Wrist}_{Robot}$} &\multicolumn{2}{c}{$T^{World}_{Robot} \cdot M^{C_1}_{World} \cdot M^{C_t}_{C_1} \cdot M^{wrist}_{C_t}$} \\ [1.5ex] 
\end{tabular}
\caption{Transformations required to calculate wrist in robot frame from passive videos to use in learning.  M denotes a transformation matrix, where T is a general transformation}.
\label{table:transformation}
\vspace{-0.14in}
\end{table}

\begin{table}[t]
\centering
\begin{tabular}{lcc}
\toprule
\textbf{Method/Task} & \textbf{Place} & \textbf{Uncover} \\
\midrule
\texttt{\ours-Noise-0.01} & 0.55 & 0.87 \\ 
\texttt{\ours-Noise-0.05} & 0.50 & 0.60 \\ 
\texttt{\ours-ImageNet} & 0.40 & 0.62 \\ 
\texttt{\ours-Transfer} &  0.60 & 0.87 \\ 
\midrule
\texttt{VideoDex-Original} & 0.70 & 0.90 \\ 
\end{tabular}
\vspace{0.05in}
\caption{\small Ablations that compare effects of different action, visual and physical priors, as well as seeing how pretraining on different data transfers to other tasks.}
\label{tab:ndp-abl}
\end{table}

\section{Retargeting Details}

We first retarget human videos from Epic-Kitchens \cite{EPICKITCHENS}. Specifically, we use the new data (refresher) from their GoPro Hero 7 Black. We retarget video clips of humans completing tasks that are similar to the robot task. These clips are on average 5-10 seconds each, depending on the task.

\paragraph{Wrist in Camera frame} 
The goal of Perspective-n-Point is to estimate the pose of the calibrated camera given a set of N 3D points in the world and their corresponding 2D point projections in the image.  First the camera must be calibrated.  To do this, we use COLMAP \cite{schoenberger2016mvs} on a set of videos.  It tracks keypoints through frames and estimates the calibration from the internet videos.   We find these camera intrinsics for the GoPro:

$$
\begin{bmatrix}
 2304.002572862 & 0 & 960 \\ 
 0 & 2304.002572862 & 540 \\  
 0 & 0 & 1  
\end{bmatrix}
$$

Using this calibration, we can now complete the Perspective-n-Point process. We are given two sets of points, 16 points in 3D on the hand model in the model's frame $\left[\begin{array}{c c c c}
X_{w}, 
Y_{w}, 
Z_{w}, 
1
\end{array}\right]^t$, and another set of 16 2D points in image frame $\left[\begin{array}{c c c}
u,
v,
1,
\end{array}\right]^t$:
$$
\left[\begin{array}{c}
u \\
v \\
1
\end{array}\right]=\left[\begin{array}{ccc}
f_{x} & 0 & c_{x} \\
0 & f_{y} & c_{y} \\
0 & 0 & 1
\end{array}\right]\left[\begin{array}{cccc}
1 & 0 & 0 & 0 \\
0 & 1 & 0 & 0 \\
0 & 0 & 1 & 0
\end{array}\right]\left[\begin{array}{cccc}
r_{11} & r_{12} & r_{13} & t_{x} \\
r_{21} & r_{22} & r_{23} & t_{y} \\
r_{31} & r_{32} & r_{33} & t_{z} \\
0 & 0 & 0 & 1
\end{array}\right]\left[\begin{array}{c}
X_{w} \\
Y_{w} \\
Z_{w} \\
1
\end{array}\right]
$$
We use the OpenCV3 solvePnPRANSAC to complete this calculation.  This implementation ensures that the process is resilient to erroneous detections.

\paragraph{Camera in First Camera frame}
In the SLAM section, the goal is to track the camera through the video. This is required to compensate for the movement of the camera.  We use this on a selection of videos where we found the camera to move significantly.

We start the SLAM process two seconds before the action clip begins.  We mark the start of the action's frame as the first frame, run it through SLAM and then recover the trajectory of the camera through the entire clip.  Specifically we recover the transformation: $M^{C_t}_{C_1}$

The process of monocular SLAM is only valid up to a scale factor. Although Epic Kitchens has noisy accelerometer and gyro information from the camera's sensors, we do not use this data to disambiguate this scale factor throughout the duration of the video clip.

ORBSLAM3 \cite{ORBSLAM3TRO} only evaluates real-time video going forward through time and does not recalculate prior poses from future information.  While this seems imperfect for this purpose, we find that the results are satisfactory for our purpose.  In our setup, we are using these retargeted videos as a prior for learning.  These retargeted trajectories are not used directly on the robot so they do not need complete accuracy.  The more important characteristic is speed. COLMAP \cite{schoenberger2016mvs} can take hours to process larger video clips, but ORBSLAM3 \cite{ORBSLAM3TRO}  can complete this process faster than real time.  This enables us to use many videos as an action prior for the robot behavior.  

\paragraph{Camera Parallel to Ground}

Now that we have the trajectory in the $C_1$ frame after SLAM and PnP, we still are missing some key transformations to get into the robot frame.  First, the $C_1$ is not always upright compared to gravity, but the robot always is.  If we have a vector normal to the ground either from the synthesized pseudo-depth map from the original Videodex method or privledged information from an accelerometer (accelerometer is not used in the original Videodex method) we can use:

\begin{equation}
\label{eq:pitch2}
\text{pitch} = \tan^{-1}(x_{Acc}/ \sqrt{y_{Acc}^2 + z_{Acc}^2})
\end{equation}
\vspace{-0.15in}
\begin{equation}
\label{eq:roll2}
\text{roll} = \tan^{-1}(y_{Acc}/ \sqrt{x_{Acc}^2 + z_{Acc}^2})
\end{equation}
\begin{equation}
\label{eq:theta2}
\text{theta} = \tan^{-1}(\sqrt{x_{Acc}^2 + y_{Acc}^2} / z_{Acc})
\end{equation}

\begin{table}[h!]
\begin{center}
    \begin{tabular}{lcc}
        \toprule
        Task & Robot Demos & Objects\\
        \midrule
        Pick& 125 & 8 \\ [1.5ex]
        Rotate& 140 & 8 \\ [1.5ex] 
        Open& 120 & 4 \\[1.5ex]
        Cover & 124 & 12  \\[1.5ex]
        Uncover& 145 & 12  \\[1.5ex]
        Place& 175 & 10 \\[1.5ex]
        Push& 136 & 14 \\[1.5ex]
        \end{tabular}
        \caption{Left: Number of trajectories we used for each task.  Robot data is collected locally using teleportation.  Most of these trajectories are 5-15 seconds in length and capture the motion trajectory of the task and visual data.  Right:  The number of different objects we used for each task's data collection.  In our testing, we show generalization outside of this set of objects.}
\end{center}
\vspace{-0.2in}
\end{table}

The pitch and roll would be used to make the trajectory upright.  The yaw is not something that is calculable this way because this rotation is around the z axis, or the direction of gravity so it isn't detected by an accelerometer.  The theta represents how far the accelerometer z axis is off from upright but is not useful to reorient the frame.

\paragraph{Accelerometer Robot Reorientation}
There's book-keeping transformations that must be included to rotate everything into the same frame conventions.  Accelerometers, like the one in the GoPro have their frame where Z is up, y is into the screen from the lens, and x is to the left if you're looking at the screen. The camera frame has the x-axis pointing to the right from the screen point of view, the y-axis facing down, and the z-axis facing out of the lens. The robot frame has its x-axis facing out towards the table, the y-axis faces to the left from the robot point of view, and the z-axis points up.  
This then leads to the following results.  The camera in world frame in roll, pitch, yaw using fixed axis is: $[pitch, 0, -roll]$.  The world frame to robot frame rotation in roll, pitch, yaw using fixed axis is $[-3.14/2,0 ,-3.14/2]$  This is used to rotate the trajectories to the robot frame and is the rotation component of $T^{World}_{Robot}$.

\paragraph{Rescaling for Robot}
We must fit the trajectories from the human videos into the robot frame.  The robot frame has significant workspace limits that the human does not have.  Even if the human arm is smaller than the robot's, the human can walk around whereas the robot arm cannot move from the middle of the table.  We therefore center the trajectory and ensure it fits in the robot frame. This is the scaling portion of $T^{World}_{Robot}$.

We rescale each dimension of the arm trajectory as:

$$M^{Wrist_N}_{World} = M^{Wrist_N}_{World} - (\max(M^{Wrist_1..N}_{World}) + \min(M^{Wrist_1..N}_{World}))/2 + \text{robotWorkspaceCenter}$$

 We would like to generate more similar trajectories to use in possible data augmentation.  The naive method is to add gaussian noise to the trajectory.  While this can be valid, it adds noise to an already noisy system.  Instead we leverage the coordinate frames to create more accurate trajectories.  We randomize the workspace scaling that is used by 10 percent.  Additionally, we create a rotation $M^{World}_{World}$ that rotates the initial world frame by up to 10 degrees in each fixed axis in roll pitch yaw convention.  This perturbs the direction that the robot moves in its frame.

While this augmentation can be helpful with lower amounts of internet data, in our results it was not used as it led to similar results to not using data augmentation.

We interpolate the length of the trajectories using RBF basis functions.  All trajectories from the internet data are rescaled to 200 datapoints. This uniformity enables efficient batch training and was used for all of the results.

\paragraph{Hand Re-targeting} 

We use a similar approach to retargerting as \citet{robo-telekinesis} and  \citet{handa2020dexpilot}. Specifically, we use the the detected human hand poses using MANO \citet{MANO:SIGGRAPHASIA:2017} (and FrankMocap \citet{FrankMocap_2021_ICCV}) to match 3D keypoints between human hands and the allegro hand. Given human hand parameters $(\beta, \theta)$, the goal is to minimize the difference between human and robot keypoints:  Human $v_i^h$ and robot $v_i^r$. The robot keypoints are a function of robot joint pose: $q$. This is done by the implicit energy function ($c_i$ are scale hyperparameters): 

\begin{equation}
    E_{\pi}( \ (\beta_{h}, \theta_{h}), \  q \ ) = \sum_i || v_i^h- (c_i \cdot v_i^r) ||_2 ^2
    \label{eq:energy2}
\end{equation}

This is inefficient to compute in real time, thus similarly to \citet{robo-telekinesis}, we distill this into a single neural network $$f_\text{hand}((\beta_{h}, \theta_{h})) = \hat{q}$$

This network learns to minimize the energy function $E_{\pi}$ described above and is trained by observing internet videos. The hand retargeting setup can be seen in Figures 3 and 4 of the main paper.

\subsection*{Task}
\begin{figure}[t]
 \vspace{-0.1in}
 \centering
 \includegraphics[scale=0.45]{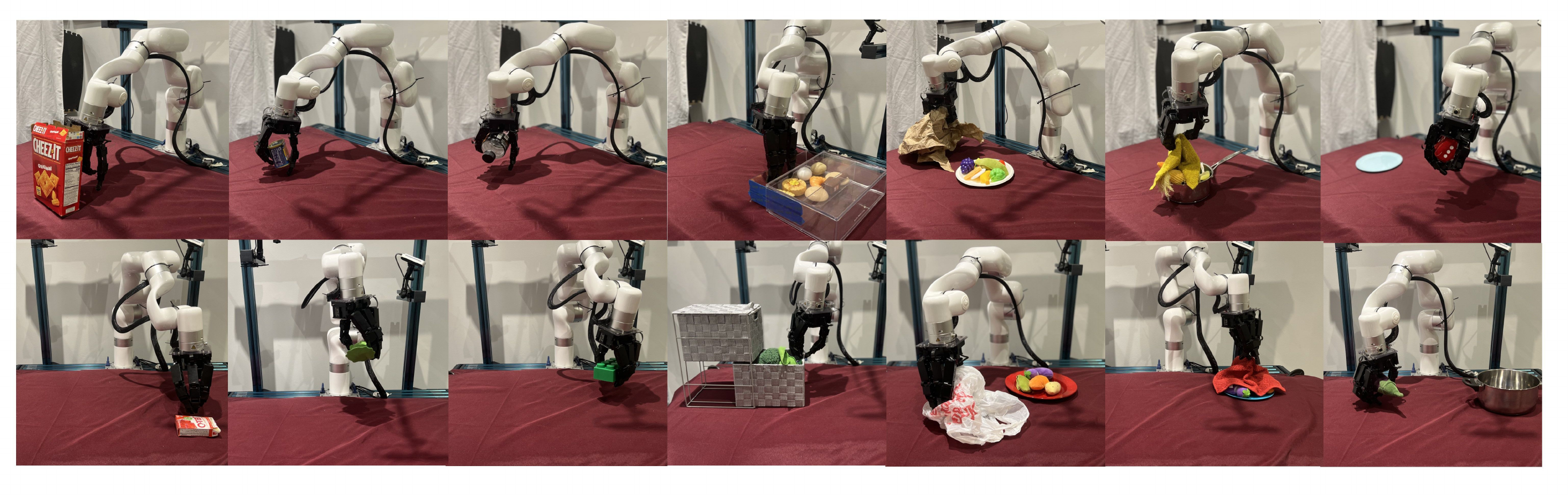}
 \vspace{-0.1in}
\caption{\small \textbf{Task Images}. A more detailed look at the tasks completed by \ours: push, pick, rotate, open, cover, uncover and place.  and our website at \url{https://videodex.github.io} for further details.}
 \label{fig:task_appendix}
\end{figure}

The tasks that were completed are Pick, Rotate, Open, Cover, Uncover, Place, Push. In pick, the task is to pickup objects off the table, or a plate/pan. Rotate involves turning an object in place. Open involves opening a drawer. Cover and uncover involve putting on or removing some form of cloth (dish, rubber, paper or plastic) on or from a plate. For place, the robot has to pickup an object and drop it in a plate or pot/pan. For push, the robot has to poke the object with its fingers. These tasks can be seen in Figure~\ref{fig:task_appendix}. We used videos from Epic Kitchens \cite{EPICKITCHENS} that were as close as possible to these tasks, and doing similar types of actions. More details can be found in Table~\ref{tab:hyperparam}.

\begin{table}[t]
\centering
\resizebox{\linewidth}{!}{%
\begin{tabular}{lcccccccccccccc}
\toprule
& \multicolumn{2}{c}{Pick} & \multicolumn{2}{c}{Rotate} & \multicolumn{2}{c}{Open} & \multicolumn{2}{c}{Cover}  & \multicolumn{2}{c}{Uncover}  & \multicolumn{2}{c}{Place}  & \multicolumn{2}{c}{Push} \\ 
& train & test & train & test & train & test & train & test & train & test  & train & test  & train & test \\
\midrule
\texttt{BC-NDP} \citep{bahl2021hndp} & 0.64 $\pm$ 0.11 & 0.38 $\pm$ 0.13 & 0.94 $\pm$ 0.06 & 0.56 $\pm$ 0.13 & 0.90 $\pm$ 0.10 & 0.60 $\pm$ 0.16 & 0.78 $\pm$ 0.15 & 0.58 $\pm$ 0.15 & 0.88 $\pm$ 0.13 & 0.82 $\pm$ 0.12 & 0.70 $\pm$ 0.15 & 0.35 $\pm$ 0.11 & 1.00 $\pm$ 0.00 & 0.71 $\pm$ 0.13 \\
\texttt{BC-Open}\citep{dasari2021rb2}& 0.50 $\pm$ 0.12 & 0.44 $\pm$ 0.13 & 0.72 $\pm$ 0.11 & 0.38 $\pm$ 0.13 & 0.80 $\pm$ 0.13 & 0.40 $\pm$ 0.16 & 0.44 $\pm$ 0.18 & 0.58 $\pm$ 0.15 & 1.00 $\pm$ 0.00 & 0.91 $\pm$ 0.09 & 0.40 $\pm$ 0.16 & 0.25 $\pm$ 0.10 & 1.00 $\pm$ 0.00 & 0.93 $\pm$ 0.07 \\
\texttt{BC-RNN} \citep{dasari2021rb2}& 0.56 $\pm$ 0.12 & 0.31 $\pm$ 0.12 & 0.78 $\pm$ 0.10 & 0.50 $\pm$ 0.13 & 0.90 $\pm$ 0.10 & 0.50 $\pm$ 0.17 & 0.56 $\pm$ 0.18 & 0.42 $\pm$ 0.15 & 0.88 $\pm$ 0.13 & 0.75 $\pm$ 0.13 & 0.70 $\pm$ 0.15 & 0.50 $\pm$ 0.11 & 1.00 $\pm$ 0.00 & 1.00 $\pm$ 0.00\\
\midrule
\textbf{\texttt{\ours}} &              0.81 $\pm$ 0.09 & 0.75 $\pm$ 0.11 & 0.89 $\pm$ 0.08 & 0.69 $\pm$ 0.12 & 0.90 $\pm$ 0.10 & 0.80 $\pm$ 0.13 & 0.78 $\pm$ 0.15 & 0.67 $\pm$ 0.14 & 1.00 $\pm$ 0.00 & 0.90 $\pm$ 0.10 & 0.90 $\pm$ 0.10 & 0.70 $\pm$ 0.11 & 1.00 $\pm$ 0.00 & 1.00 $\pm$ 0.00 \\
\bottomrule
\end{tabular}}
\vspace{0.05in}
\caption{\small  We present the variance of train objects and test objects for Videodex and baselines described above.  See the main paper for the mean results.}
\vspace{-0.2in}
\label{tab:var}
\end{table}

\section{Learning Pipeline Details}

\paragraph{Learning Setup} For our approach, we use the ResNet18 from R3M \cite{r3m} weights as the visual backbone. This produces an intermediate feature vector of size 512. This is processed with a 2 layer MLP with a hidden dimension of 512. The visual features are concatenated with the starting hand and wrist pose. We employ two such MLPs, one for the hand and wrist trajectories. These are then processed with an NDP \cite{bahl2020neural}. The NDP processes the input with a single hidden layer to project it into the desired size (parameters $W$ and $g$). For more information we point the readers to \citet{bahl2020neural}. We use the implementation from \citet{dasari2021rb2}. We use standard data augmentations from Pytorch.  Specifically, we use RandomResizeCrop from a scale of 0.8 to 1.0.  We use RandomGrayscale with a probability of 0.05  We use ColorJitter with a brightness of 0.4, contrast of 0.3, saturation of 0.3 and hue of 0.3.  Finally, we normalize the RGB values around the typical mean and standard deviation for color images:  $\mu = (0.485, 0.456, 0.406)$  $\sigma = (0.229, 0.224, 0.225)$.  For different baselines, we used the same backbone (R3M \cite{r3m}) as our method. We use the same architecture style as well, with the visual features being processed by both a wrist and hand stream. Finally, all network sizes are the same or very similar. We describe our hyperparameters in Table~\ref{tab:hyperparam}. 

\begin{table}[ht]
\centering
\begin{tabular}{lc}
\toprule
\textbf{Parameter} & \textbf{Value}\\
\midrule
Learning Rate &$1 \times 10^{-3}$ \\
Batch Size & 32 \\ 
Training Demonstrations Per Task & 120-175 \\
Human Videos Per Task & 350 (Cover/Unicver, Rotate, Push) - 2500 (Open, Pick, Place) \\
Trajectory Length Human Videos & 200 (rescaled) \\ 
NDP \cite{bahl2020neural} Basis Functions $N$ & 300 \\ 
NDP \cite{bahl2020neural} Global Parameter $\alpha$ & 15 \\ 
\bottomrule
\end{tabular}
\vspace{0.1cm}
\caption{Parameter List}
\label{tab:hyperparam}
\end{table}

\section{Experimental Setup}
We collect data using a dexterous hand robotic teloperation setup \cite{robo-telekinesis, handa2020dexpilot}. A trained operator stands in front of the camera within view of the robot and operates the system in real-time to collect demonstrations with a trained, uniform style.  Another manager stands by.  The goal of this manager is to place items on the table for manipulation, randomize locations and types of objects, to start and stop demonstrations for the operator and manage the robot system. We collect about 120-175 demonstrations per task.  See table \ref{table:transformation} for details.

\section{Hardware Details}

Our hardware setup consists of an LEAP 16 DOF Hand and an XArm 6 manipulator (from Ufactory). The hand is mounted on the wrist of the XArm. To collect data we use a similar teleoperation system as provided by \citet{robo-telekinesis} and \citet{handa2020dexpilot}. We use Intel Realsense D415 cameras to collect human teleoperation and robot videos. We use four NVIDIA RTX 2080TI's for training the policy and running the teleoperated system. 
We experiment with the Allegro Hand and use it to collect some teleoperated demonstration data, but we find it to be very unreliable and break many times.  The motors also quickly overheat and are weak in practice.  Therefore, to alleviate these issues we use the LEAP Hand for collecting the final results.

\subsection*{External Codebases}

We use the following different external codebases for our pipeline: 

\begin{itemize}

    \item Human body and hand detection: FrankMocap \cite{FrankMocap_2021_ICCV}, \url{https://github.com/facebookresearch/frankmocap}
    \item NDP and Behavior Cloning \cite{dasari2021rb2} code from \url{https://github.com/AGI-Labs/robot_baselines} 
    \item Code for R3M \cite{r3m} \url{https://github.com/facebookresearch/r3m}
    \item CQL baseline from \citet{seno2021d3rlpy} (\url{https://github.com/takuseno/d3rlpy})
    \item COLMAP from \citet{schoenberger2016mvs} (\url{https://github.com/colmap/colmap})
    \item ORBSLAM3 from \citet{ORBSLAM3TRO} (\url{https://github.com/UZ-SLAMLab/ORB_SLAM3})
    \item GoPro Metadata Extractor from (\url{https://github.com/JuanIrache/gpmf-extract})
    \item Rigid Transform class from (\url{https://github.com/BerkeleyAutomation/autolab_core/blob/master/autolab_core/rigid_transformations.py})
    \item PnP from (\url{https://opencv.org/})
\end{itemize}

\end{document}